\documentclass[lettersize,journal]{IEEEtran}
\usepackage{amsmath,amsfonts}
\usepackage{algorithmic}
\usepackage{algorithm}
\usepackage{array}
\usepackage{stfloats}
\usepackage{url}
\usepackage{verbatim}
\usepackage{textcomp}
\usepackage{soul}
\usepackage{graphicx}
\usepackage{cite}
\usepackage{xcolor}
\usepackage{multirow}
\usepackage{pifont}
\usepackage{diagbox}
\usepackage{mathabx}
\usepackage{subcaption}
\def\BibTeX{{\rm B\kern-.05em{\sc i\kern-.025em b}\kern-.08em
    T\kern-.1667em\lower.7ex\hbox{E}\kern-.125emX}}
\usepackage{balance}

\begin{document}

\title{An Identity-Preserved Framework for Human Motion Transfer}

\author{Jingzhe Ma, Xiaoqing Zhang, and Shiqi Yu{$^*$},~\IEEEmembership{Member,~IEEE},
\thanks{The authors are with the Department of Computer Science and Engineering, Southern University of Science and Technology, Shenzhen 518055, China}
\thanks{Shiqi~Yu is the corresponding author (email: yusq@sustech.edu.cn).

}
}


\markboth{An Identity-Preserved Framework for Human Motion Transfer}%
{Shell \MakeLowercase{\textit{et al.}}: A Sample Article Using IEEEtran.cls for IEEE Journals}


\maketitle

\begin{abstract}
 
Human motion transfer (HMT) aims to generate a video clip for the target subject by imitating the source subject's motion.
Although previous methods have achieved good results in synthesizing good-quality videos, they lose sight of individualized motion information from the source and target motions, which is significant for the realism of the motion in the generated video.
To address this problem, we propose a novel identity-preserved HMT network, termed \textit{IDPres}.
This network is a skeleton-based approach that uniquely incorporates the target's individualized motion and skeleton information to augment identity representations. 
This integration significantly enhances the realism of movements in the generated videos. 
Our method focuses on the fine-grained disentanglement and synthesis of motion.
To improve the representation learning capability in latent space and facilitate the training of \textit{IDPres}, we introduce three training schemes.
These schemes enable \textit{IDPres} to concurrently disentangle different representations and accurately control them, ensuring the synthesis of ideal motions.
To evaluate the proportion of individualized motion information in the generated video, we are the first to introduce a new quantitative metric called Identity Score (\textit{ID-Score}), motivated by the success of gait recognition methods in capturing identity information.
Moreover, we collect an identity-motion paired dataset, $Dancer101$, consisting of solo-dance videos of 101 subjects from the public domain, providing a benchmark to prompt the development of HMT methods.
Extensive experiments demonstrate that the proposed \textit{IDPres} method surpasses existing state-of-the-art techniques in terms of reconstruction accuracy, realistic motion, and identity preservation.
\end{abstract}

\begin{IEEEkeywords}
human motion transfer, disentanglement representation, gait recognition, video generation.
\end{IEEEkeywords}

\section{Introduction} \label{intorduction}
Imitating human motion holds significant importance across various domains, such as augmented reality (AR)~\cite{kim2016retargeting}, virtual reality (VR)~\cite{slater2010first}, movie production~\cite{korban2022survey}, robotics~\cite{ayusawa2017motion}, and computer vision~\cite{yang2020transmomo, sun2022human}. 
With the recent surge of development in Generative Adversarial Nets (GANs)~\cite{goodfellow2020generative}, the human motion transfer (HMT) task has become a popular way to obtain the desired human motion imitation.
Specifically, HMT aims to generate a video of one person (Subject B), and this person performs the same motion as another person (Subject A).

\begin{figure}[t]
    \centering
        \includegraphics[width=\linewidth]{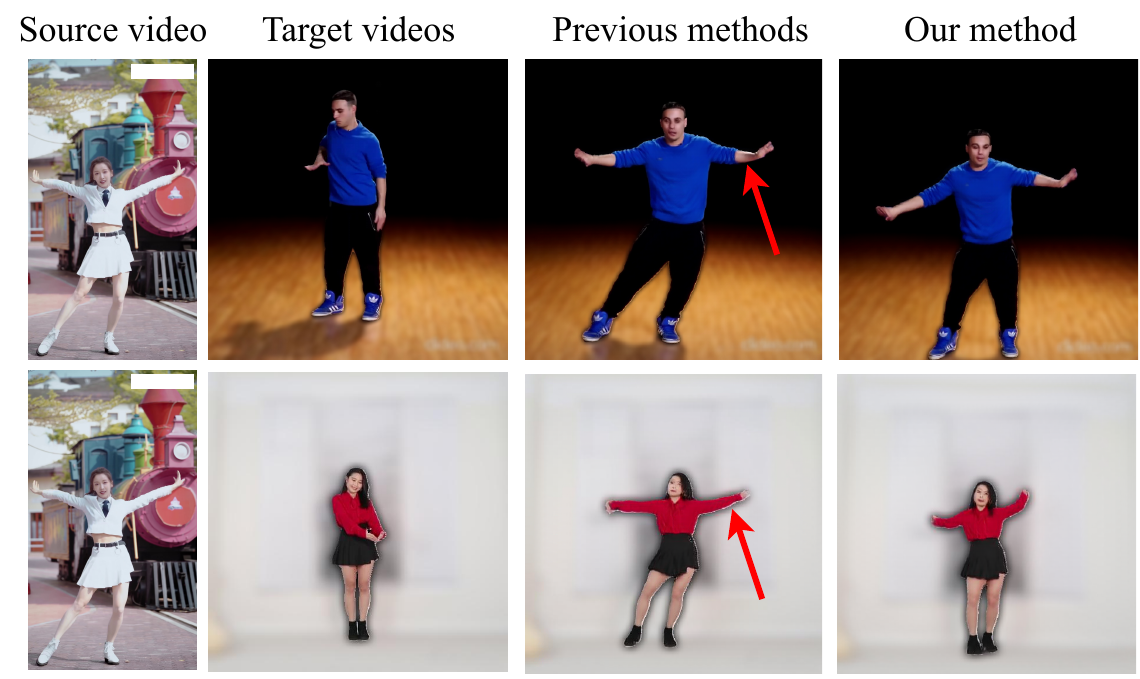}
    \caption{An example of an unnatural case is presented by TransMoMo \cite{yang2020transmomo}.
    The motion from the source video (the girl in white) is transferred to the target videos (the man in blue, and the girl in red). The synthesized videos from previous methods show a limitation: regardless of the subjects' identities, they tend to exhibit similar postures. In contrast, our proposed method considers individualized information and generates the natural posture of each individual.}
    \label{fig:challenges}
\end{figure}

Most HMT methods firstly decompose the motion and appearance of two subjects and then recompose the source subject's motion with the target subject's appearance~\cite{kappel2021high,tao2022structure, ni2022cross, gomes2021shape, sun2022human,ren2020deep,huang2021few, wang2019few, yang2020transmomo}.
Despite the fact that these methods have achieved remarkable results in synthesizing good-quality videos, a limitation is their tendency to generate videos with unnatural motion.
An example is shown in Fig.~\ref{fig:challenges}, where the motion of a girl in white (source) is transferred to two different individuals (a man in blue and a girl in red). 
The synthesized videos show that both the man in blue and the girl in red perform motions similar to those of the source video.
Nevertheless, the same motion performed by two individuals in the real world cannot be exactly the same in all details.
This is because each individual has his/her own personal motion style.
Those styles can be viewed as individualized motion information in human motion.
Existing HMT methods often omit individualized motion information, leading to incomplete disentanglement for human motion and suboptimal identity preservation.

In this paper, we suggest that human motion should be further decomposed into coarse-grained motion content ($MC$) and fine-grained individualized motion ($IM$).
The $MC$ dictates the general type of movements, like raising hands or kicking, while the $IM$ refers to the nuanced motion of an individual, like the personal way each person might raise their hand.
Therefore, we consider the HMT task as creating a video that not only transfers the source's $MC$ to the target but also preserves the target's $IM$ and appearance.
In this context, we specifically refer to appearance as human structure ($HS$), which is a static feature, \textit{i.e.} does not change with motion.

\begin{figure*}[t]
	\centering
    \includegraphics[width=1\linewidth]{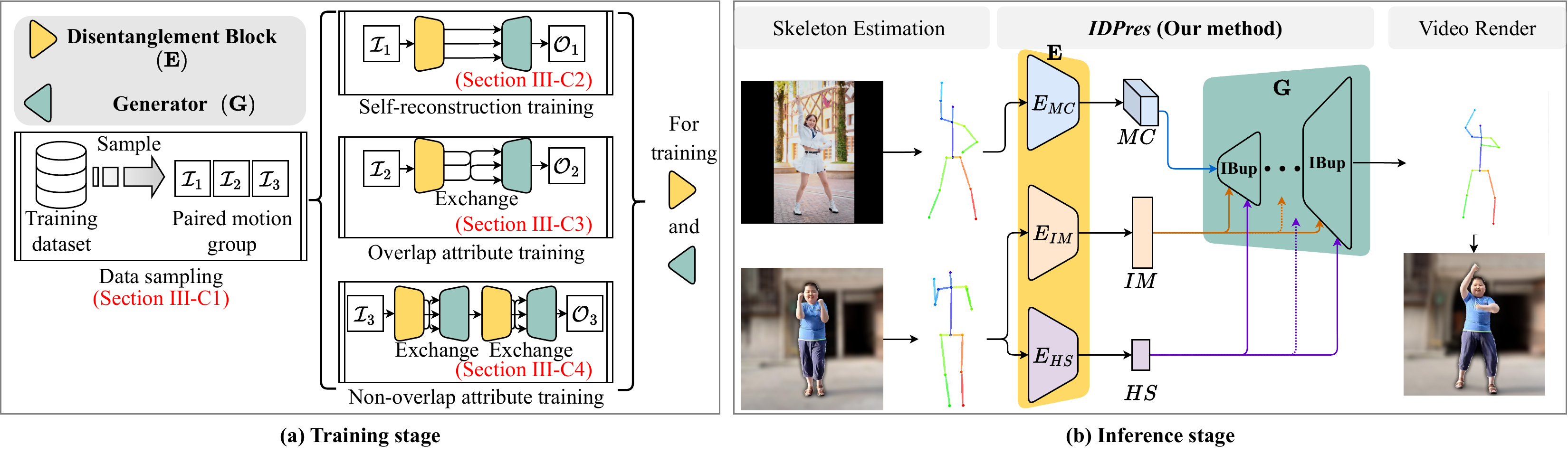}
	\caption{The overview of the proposed \textit{IDPres}. (a) The overview of the training stage. Our training stage contains specific data sampling (Section~\ref{subsubsec:data_sampling}) and three training schemes: self-reconstruction training (Section~\ref{subsubsec:self_reconstruction}), overlap attribute training (Section~\ref{subsubsec:overlap}), and non-overlap attribute training (Section~\ref{subsubsec:nonoverlap}). (b) The overview of the inference stage. This stage takes two videos with different subjects and movements as input.}
	\label{fig:IDpres}
\end{figure*}

In order to achieve this task, two challenges need to be addressed.
First, decomposing the fine-grained and high-frequency $IM$ information from human motion and using it to achieve human transfer is difficult.
Second, to our best knowledge, there are no available evaluation metrics for evaluating the proportion of $IM$ in the generated video.

To address the first challenge, we introduce \textit{IDPres}, an identity-preserved network. 
\textit{IDPres} employs a skeleton-based structure~\cite{yang2020transmomo} for focused motion analysis, unaffected by variations such as clothing or carrying.
\textit{IDPres} focus on leveraging the target subjects' $IM$ and skeleton information ($HS$ mentioned above) to improve the reality of motion in generated videos. 
Its methodology is detailed in Fig~\ref{fig:IDpres}. 
Moreover, \textit{IDPres} not only facilitates flexible motion transfer but also aids in deconstructing human motion components. 
Inspired by Group Supervised Learning (GSL)~\cite{ge2021zero}, we implement three training schemes to enhance \textit{IDPres}'s ability to capture $IM$ information, detailed in Section~\ref{sec:tp}.
In the inference stage, leveraging the trained \textit{IDPres}, users can compose a single harmonized motion formed by fusing three latent codes via an intuitive “plug-and-play” interface.

Recent studies have successfully demonstrated that leveraging subjects' personal style (both $IM$ and $HS$) in walking is beneficial to identifying individuals~\cite{fan2020gaitpart,fan2023exploring,fan2023opengait,chao2021gaitset}.
Gait recognition methods, based on this principle, employ machine learning methods to represent personal style during walking, thereby facilitating identity recognition despite variations such as multiple camera viewpoints, various clothing, and carried items.
They typically regard this personal style as an individual's \textbf{identity information}. 
Moreover, we experimentally found that the gait recognition methods can also recognize the subject's identity information based on their non-walking motion. 
Thus, we argue that gait recognition methods can efficiently capture identity information from dance movements by considering them as identity information extractors.
Based on this extractor, we propose a novel metric, named Identity Score (\textit{ID-Score}), for quantitatively evaluating the preservation extent of $IM$ information in generated motion sequences, aiming to handle the second challenge.


Moreover, data is essential for achieving a reasonable result of the deep neural network.
Although several publicly available datasets~\cite{mixamo, chen2015utd} can satisfy the condition that multiple subjects perform the same motion, these datasets have relatively few subjects or perform only simple movements.
Thus, we collected a dataset from the Internet, named $Dancer101$, with 101 subjects' solo-dance videos.
In the dataset, multiple subjects perform the same motion (choreography), and the motions are complex dance motions.

We summarize our \textbf{contributions} as follows:

\begin{enumerate}
    \item We introduce an identity-preserved human motion transfer framework, \textit{IDPres}, which utilizes both $IM$ and $HS$ to alleviate the phenomenon of unrealistic motion in generated videos. We also present three training schemes to promote the training of \textit{IDPres}.
    
    \item We experimentally found gait models also can capture identity information from non-walking motion. Thus, we leverage gait models as identity information extractors and propose a new metric, Identity-Score (\textit{ID-Score}), to quantitatively evaluate the preservation extent of $IM$ information in generated motion sequences.
    %
    
    \item We collected a dataset, named $Dancer101$,  for identity-preserved human motion transfer, and the dataset contains solo-dance videos from 101 subjects.
    
    \item Extensive experiments are conducted. The results demonstrate that our methods outperform the state-of-the-art in both reconstruction performance and \textit{ID-Score} on three datasets. Moreover, the visualization shows that \textit{IDPres} efficiently transfers the source motion to the target subject while preserving their $IM$ and $HS$.
\end{enumerate}

\section{Related Works}

\subsection{Human motion transfer}
Human motion transfer methods can  be roughly divided into three categories based on data modality: image-based~\cite{yu2022shature, li2019dense, siarohin2018deformable, balakrishnan2018synthesizing}, video-based~\cite{chan2019everybody, wang2018video, wang2019few, mallya2020world, huang2021few, ren2020deep, sun2022human}, and skeleton-based ones~\cite{chan2019everybody, aberman2019learning, aberman2020skeleton, yang2020transmomo, zhu2022mocanet}.

\textbf{Image-based methods} refer to generating an image by transferring the action of the source subject to the target subject.
In order to generate a video, frames are generated one by one separately, and there are no constraints among the frames.
Siarohin \textit{et al.}~\cite{siarohin2018deformable} employed a conditional generative adversarial network (cGAN)~\cite{mirza2014conditional} to transfer the source subject's posture to the target subject and then generated images by a given source skeleton map as conditions.
Balakrishnan \textit{et al.}~\cite{balakrishnan2018synthesizing} presented a modular generative neural network that synthesized unseen poses by training pairs of images and poses. Li \textit{et al.}~\cite{li2019dense} proposed a feedforward appearance flow generation module to efficiently encode the dense and intrinsic correspondences in 3D space for human pose transfer. Li \textit{et al.}'s method~\cite{li2019dense} can 
address the self-occlusion problem of skeleton key points. 
However, the performance of this method is decided by the quality of human models. 
To alleviate this problem, Ren \textit{et al.}~\cite{ren2020deep} presented a differentiable global-flow local-attention block to reassemble the inputs at the feature level for pose generation. 
Image-based methods omit temporal context, thus difficult to extract $IM$ from human movements.

\textbf{Video-based methods} can introduce temporal information that cannot be handled in image-based ones. 
Wang \textit{et al.}~\cite{wang2018video} implemented video-to-video synthesis by converting semantic videos into frames level. 
They employed an off-the-shelf algorithm, DensePose, to estimate the 3D surface of a human body as the input. Spatio-temporal generators and discriminators are used to transfer the motion for the target subject. 
Chan \textit{et al.}~\cite{chan2019everybody} used skeletons as an intermediate representation for frame-to-frame transfer. 
They considered temporal information by the two consecutive frames as the generator's input and the discriminator's input.
Meanwhile, a temporal smoothing loss function is employed to constrain the temporal representation. 
However, video-based methods still face two problems: \textit{i)} they cannot be generalized to subjects who are not in the training set. \textit{ii)} they tend to fail in generalizing long videos. 
To address the first problem, Wang \textit{et al.}~\cite{wang2019few} applied few-shot learning~\cite{wang2020generalizing} and proposed a few-shot video-to-video framework to synthesize videos of unseen subjects. 
To improve the quality of the generated videos, Huang \text{et al.}~\cite{huang2021few} rendered a human texture map to a surface geometry (represented as a UV map). 
To handle the second problem, Mallya \textit{et al.}~\cite{mallya2020world} introduced a novel framework by bolstering video-to-video models, which was achieved by condensing the 3D world rendered into a physically-grounded estimate.
However, those methods will introduce extra training parameters that may increase the difficulty of network training.

\textbf{Skeleton-based methods} are the most intuitive methods for handling temporal information. 
They can also have fewer parameters than image or video-based ones since the input skeleton data is lower dimensional than images or videos.
Skeleton-based methods normally consist of three steps: video-to-skeleton estimation, skeleton-to-skeleton transfer (motion retargeting), and skeleton-to-video~\cite{yang2020transmomo}. 
To explore human motions, skeleton-to-skeleton transfer is a mainstream research trend for skeleton-based motion transfer. 
Aberman \textit{et al.}~\cite{aberman2019learning} proposed a two-branch framework with a part confidence map as the 2D pose feature to clone the human motions in a video. 
Then, a deep neural network is trained to decompose temporal sequences of 2D poses into three components: abstract motion, skeleton, and camera view-angle. 
Recently, Yang \textit{et al.}~\cite{yang2020transmomo} proposed TransMoMo that can be trained in an unsupervised manner to decompose the aforementioned three components.
However, those methods ignore $IM$ information in human movements and will result in unrealistic motion generation.


\subsection{Disentangled representation learning}
The disentangled representation learning has obtained much attention in the image or video generation tasks~\cite{chen2016infogan, dai2021disentangling, albarracin2022video}.
As a fundamental technique in network interpretability, disentangled representation learning has been popularly adopted in human motion transfer.
It can infer latent factors from the input motion, and each latent factor is responsible for generating a semantic attribute (such as the human skeleton). 
The pioneering work by Villegas~\textit{et al.}~\cite{villegas2018neural} leveraged two RNNs to capture the motion context from source sequences and synthesize a new motion animation, respectively. 
Specifically, a reference pose of the target skeleton is provided to the RNN decoder, and a cycle consistency loss is employed to decompose the motion context and reference pose information.
Lim \textit{et al.}~\cite{lim2019pmnet} learned overall movement frame-by-frame and combined the results to construct the output sequence. 
They represented human motion as the global velocity of the root joint and relative coordinates from the root joint position. 
Kim \textit{et al.} argued that CNNs are better than RNNs for extracting motion context because they can capture short-term motions. 
Some other works~\cite{aberman2019learning, yang2020transmomo} adopted different CNNs encoders to disentangle different representations. 
Most previously mentioned methods also use a classifier or metric learning to decompose different representations. 
However, those methods cannot disentangle the $IM$ information from human movements. 
How to decompose both $IM$ and $MC$ from human movements simultaneously is still an open issue.

\subsection{Gait Recognition}

Gait recognition technologies distinguish individuals by their distinctive walking styles, and these methods are typically classified into two types: skeleton-based~\cite{liao2020model,teepe2021gaitgraph, teepe2022gaitgraph2, zhang2023spatial}, silhouette-based methods~\cite{fan2023learning, Lin2021ICCV, fan2023opengait, fan2020gaitpart, fan2023exploring}.

\textbf{Skeleton-based gait recognition} employs the intrinsic structure of the human form as input, such as the estimated 2D skeleton points, applying deep learning models to learn identity-related representations for recognition. These gait representations abstract the visual information to coordinates of human joints or vectors, aiming to reduce disturbs from other factors like clothing and carrying. PoseGait~\cite{liao2020model} is a representative work, which combines 3D skeleton data with hand-crafted characteristics to address clothing and viewpoint variations, GaitGraph~\cite{teepe2021gaitgraph, teepe2022gaitgraph2} introduces a graph convolution network for 2D skeleton-based gait representation learning, and Gait-TR~\cite{zhang2023spatial} further improves accuracy and robustness using spatial transformer networks and temporal convolution networks. Although existing methods have achieved significant progress, no previous work has attempted to apply skeleton-based methods to dance movement data.
Moreover, the skeleton estimation results of skeleton-based methods are easily affected by the environment and the camera's viewpoints, significantly impacting recognition accuracy.

\textbf{Silhouette-based gait recognition} methods mostly learn gait features from silhouette images, leveraging the potential informative visual characteristics. 
With the advent of deep learning, the focus of these methods is changed to spatial feature extraction and temporal gait modelling. Specifically, GaitSet~\cite{chao2021gaitset} treats the gait sequence as a set and employs a maximum function to compress the sequence of frame-level spatial features. GaitPart~\cite{fan2020gaitpart} delves into the intricate local details of silhouette complementation by capturing temporal dependencies through a Micro-motion Module.
GaitGL~\cite{Lin2021ICCV} introduces both global and local 3D convolution layers, addressing the need for a holistic understanding of gait without neglecting important details~\cite{Lin2021ICCV}. DeepGaitV2~\cite{fan2023exploring}, presents profound insights into deep learning models for outdoor gait recognition, showing promising performance on various datasets.
Despite the surpassing performance of silhouette-based methods across different benchmarks, they suffer challenges such as changes in clothing or carrying and varying camera viewpoints, which can distort the shape information in silhouettes and influence recognition performance.

According to the above analysis, we find that compared with skeleton-based gait recognition methods, silhouette-based methods are relatively easier to be affected by factors, such as clothing and carrying conditions. 
Considering that skeleton-based gait recognition methods emphasize intrinsic structural features, such as skeletons~\cite{liao2020model}, which tend to be more robust against various external factors. Hence, we leverage skeleton-based methods as identity information extractors. This allows us to quantitatively evaluate the extent of $IM$ information from generated motion clips.

\section{Method}\label{methods}

We begin by presenting the preprocessing details of our collected dataset, $Dancer101$ (Section~\ref{sec:data_pre}). 
Subsequently, an overview of \textit{IDpres} is provided in Fig.~\ref{fig:IDpres}.
Following this, various components within \textit{IDpres} are described, with a detailed explanation of the training scheme.
Finally, the new metric \textit{ID-Score} is described.
In particular, \textit{IDPres} consists of two blocks: a disentanglement block (formulated $\mathbf{E}$) and a generator block (formulated $\mathbf{G}$), which are described in Section~\ref{sec:dis-block} and Section~\ref{sec:generator}, respectively.
The disentanglement block contains three well-designed encoders, \textit{i.e.} a motion content aware encoder ($E_{MC}$) for capturing $MC$, an individualized-aware encoder ($E_{IM}$) for capturing $IM$, and a skeleton-aware encoder ($E_{HS}$) for capturing $HS$. 
The generator block consists of several well-designed identity-broadcasted upsampling blocks (\textit{IBup}). 
\textit{IBup} can achieve an accurate fusion for $MC$, $IM$ and $HS$.
To further model long-range dependencies, identity information (both $IM$ and $HS$) can be adequately fused to the receptive field of different scales.
In Section~\ref{sec:tp}, we introduce three training schemes, self-reconstruction training, overlap attribute training, and non-overlap attribute training, all promoting disentanglement and generation for \textit{IDPres} network.
Finally, we describe the detailed design of \textit{ID-Score} in Section~\ref{sec:idscore}.

\begin{table*}[t]
\centering
\caption{Comparisons between two widely-used motion datasets and our collected datasets $Dancer101$ and $Dancer99$.}
\label{tab:dataset}
\resizebox{\textwidth}{!}{
\begin{tabular}{c|c|c|c|c|c|c|c} 
\hline
\multirow{2}{*}{Dataset}                                                               & \multicolumn{2}{c|}{\begin{tabular}[c]{@{}c@{}}Subject number\\($ID$ labels)\end{tabular}} & \multicolumn{2}{c|}{\begin{tabular}[c]{@{}c@{}}Number of motion \\ clips per subject\\($MC$ labels)\end{tabular}} & Resolution                          & \multirow{2}{*}{Collect manner} & \multirow{2}{*}{\begin{tabular}[c]{@{}c@{}}Complexity \\ of motion\end{tabular}}  \\ 
\cline{2-6}
                                                                                       & training                                                             & testing               & training       & testing                                                                                            & $W \times H$                        &                                 &                                                                                   \\ 
\hline\hline
\multirow{2}{*}{Mixamo}                                                                & 32                                                                   & 4                     & 799            & 64                                                                                                 & \multirow{2}{*}{-}                  & \multirow{2}{*}{animation}      & \multirow{2}{*}{complex motion}                                                   \\
                                                                                       & (\# 0 $-$ \# 31)                                                        & (\# 32 $-$ \# 35)         & (\# 0 $-$ \# 798) & (\# 0 $-$ \# 63)                                                                                      &                                     &                                 &                                                                                   \\ 
\hline
\multirow{2}{*}{UTD-MHAD}                                                              & 6                                                                    & 2                     & 27             & 27                                                                                                 & \multirow{2}{*}{640 $\times$ 480}   & \multirow{2}{*}{laboratory}     & \multirow{2}{*}{basic motion}                                                     \\
                                                                                       & (\# 0 $-$ \# 5)                                                          & (\# 6 $-$ \# 7)           & (\# 0 $-$ \# 26)  & (\# 0 $-$ \# 26)                                                                                      &                                     &                                 &                                                                                   \\ 
\hline
\multirow{2}{*}{\begin{tabular}[c]{@{}c@{}}$Dancer101$\\ (Our collected)\end{tabular}} & 91                                                                   & 10                    & 81             & 81                                                                                                 & \multirow{2}{*}{1080 $\times$ 1920} & \multirow{2}{*}{open domain}    & \multirow{2}{*}{complex motion}                                                   \\
                                                                                       & \begin{tabular}[c]{@{}c@{}}(\# 0 $-$ \# 69,\\\# 80 $-$ \# 100)\end{tabular} & (\# 70 $-$ \#79)         & (\# 0 $-$ \# 80)  & (\# 0 $-$ \# 80)                                                                                      &                                     &                                 &                                                                                   \\ 
\hline
\multirow{2}{*}{\begin{tabular}[c]{@{}c@{}}$Dancer99$\\ (Our collected)\end{tabular}}  & -                                                                    & 99                    & -              & 32                                                                                                 & \multirow{2}{*}{1080 $\times$ 1920} & \multirow{2}{*}{open domain}    & \multirow{2}{*}{complex motion}                                                   \\
                                                                                       & -                                                                    & (\# 0 $-$ \# 98)          & -              & (\# 0 $-$ \# 31)                                                                                      &                                     &                                 &                                                                                   \\
\hline
\end{tabular}
}
\end{table*}

\subsection{Data preparation and preprocessing} \label{sec:data_pre}

Collecting a dance dataset with paired motion and identity is crucial for disentangling $IM$ from movements.
Although open-source datasets like Mixamo~\cite{mixamo} and UTD-MHAD~\cite{chen2015utd} offer motion-identity pairing, they are constrained by animated environments or simplistic movements.
To overcome these limitations, we collect a new dataset, named $Dancer101$, consisting of solo dance videos from 101 individuals performing complex and identical dance routines. 
Each video, lasting approximately three minutes and forty seconds, presents identical choreography and background music, captured in various real-world settings with a resolution of $1080 \times 1920$ pixels. 
This dataset provides an excellent opportunity to study $IM$ information in a disentangled manner.
The processing of $Dancer101$ involved several steps:

\textbf{Video rescaling and movement alignment.} 
The videos are rescaled to a uniform resolution of $512 \times 512$  pixels, maintaining the original body proportions. 
Subsequently, we manually synchronize the subjects’ movements using their audio tracks, which is thanks to all the dance movements being driven by the same background music. 
Despite temporal alignment being imperfect, with errors within a five-frame range, this process is beneficial in constructing a dataset that pairs identity and motion effectively, facilitating the study of $IM$ information.

\textbf{Acquisition of motion clips.}
Utilizing OpenPose~\cite{openpose}, a pose estimation algorithm, we extract 2D human skeleton points from the videos, forming motion sequences. 
In our paper, we focus on the first fifteen points of the BODY\_25 format. 
In particular, the points are nose (0), neck (1), right shoulder (2), right elbow (3), right wrist (4), left shoulder (5), left elbow (6), left wrist (7), mid hip (8), right hip (9), right knee (10), right ankle (11), left hip (12), left knee (13), and left ankle (14).
Considering the complex nature of the dance movements in $Dancer101$, samples of missing points in the skeleton maps are inevitable. 
To mitigate this, frames with more than one-third missing points are discarded. 
For frames with partial missing points, we interpolate these missing points using average values from five adjacent frames.
We then segment these sequences into 81 non-overlapping motion clips per subject, each comprising 64 consecutive frames.

\textbf{Annotation and Dataset Division.}
The 8,181 motion clips ($101 \times 81 = 8,181$), obtained from the 101 subjects, are annotated with sequential numbers, \textit{i.e.} $ID$ \# 0 to $ID$ \# 100.
Similar to identity annotation, we annotate motion clips per subject into 81 different classes ($MC$ \# 0 to $MC$ \# 80).
We divide these clips into training and testing sets based on the subject level, ensuring no overlap.
The training set has 91 subjects with 7,371 ($91 \times 81 = 7,371$) motion clips, and the testing set has 10 subjects with 810 ($10 \times 81 = 810$) motion clips. 
Further, for $IM$ information evaluation using our proposed \textit{ID-Score} metric, the testing set is split into gallery and probe sets, where the former's identity labels are known, and the latter's require prediction. We allocate 40\% of each subject's motion clips to the gallery set and the remainder to the probe set. 
The training set is utilized to train both the gait recognition models and our proposed \textit{IDPres}.

To effectively evaluate the cross-domain capabilities inherent in various HMT methods, we collect an additional dataset of 99 solo dance videos, referred to as $Dancer99$. 
These videos also have the same choreography and background music, but both the choreography and background music differ from $Dancer101$. 
We leverage the same preprocessing protocols used for $Dancer101$ to prepare $Dancer99$, yielding 3,168 motion clips, with each subject contributing 32 clips. 
The detailed comparison and division of the datasets are presented in Table~\ref{tab:dataset}.


\subsection{\textit{IDPres} network architecture} 

\subsubsection{Disentanglement block} \label{sec:dis-block}
\begin{figure*}[b]
    \centering
    \includegraphics[width=0.9\linewidth]{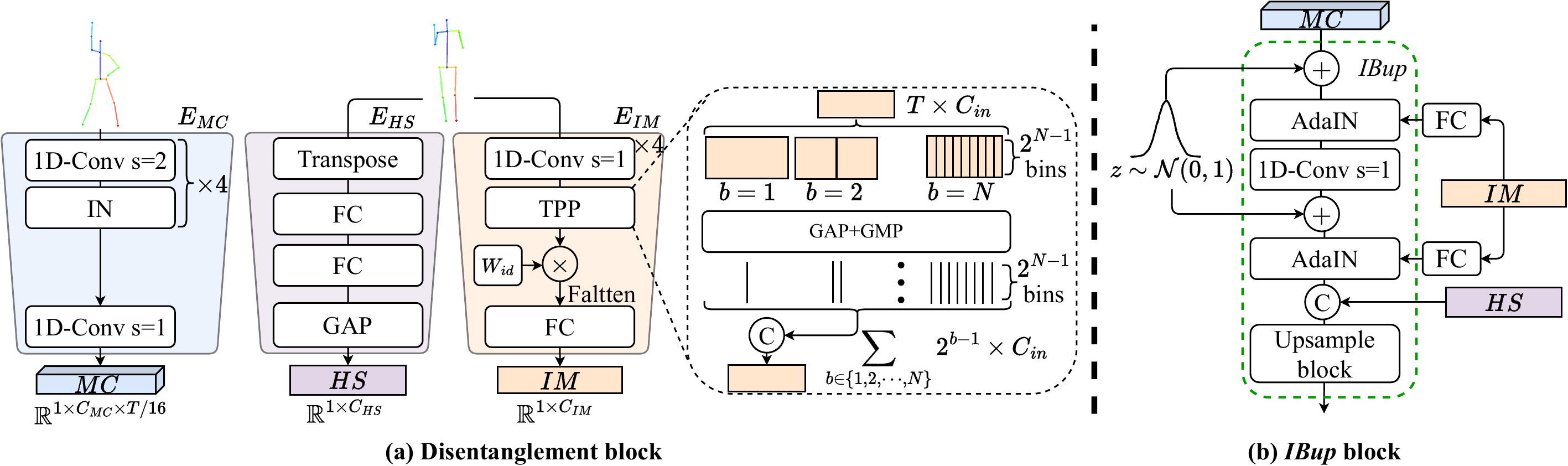}
    \caption{The overview of (a) disentanglement block, $\mathbf{E}$ and (b) identity-broadcasted upsampling blocks, \textit{IBup}. In subfigure (a), 1D-Conv s=$i$ represents the temporal convolutional layer with a stride of $i$ and a LeakyReLU activation function, IN means instance normalization, $T$ is the number of frames, $C_{k} (k \in \{MC, IM, HS\}$) is the number of channels of $k$, GAP and GMP are temporal-level global average pooling and global max pooling respectively. FC means the full-connection layer. In subfigure (b), FC is used to learn the mean and variance of $IM$, $z$ is a Gaussian noise to improve the diversity of generated motions.}
    \label{fig:dis_generator_block}
\end{figure*}

The disentanglement block, termed $\textbf{E}$, is designed to separate three different representation attributes from the input data. One of the unique advantages of our proposed method is the individualized-aware representation disentanglement, which can be used for identity-preserved human motion transfer.
As illustrated in Fig.~\ref{fig:dis_generator_block}, our disentanglement block contains three encoders of different architectures. The three encoders are the motion content aware encoder ($E_{MC}$) for capturing $MC$, the individualized-aware encoder ($E_{IM}$) for capturing $IM$, and the skeleton-aware encoder ($E_{HS}$) for capturing $HS$.

As discussed above, the $E_{MC}$ mainly extracts motion content from human movements.
We employ several temporal convolution layers in $E_{MC}$ to capture time-sensitive representations. 
Then, the instance normalization (IN)~\cite{adain} layer is applied after each temporal convolution layer to control local statistics in human movements, in other words, $IM$ information.
Benefiting from the IN layer, coarse and low-frequency details in the temporal dimension are captured (\textit{i.e.} global motion).

In the $E_{IM}$, unlike $E_{MC}$, the IN layer is not applied since the $IM$ information needs to be preserved. 
The $IM$ representation is one of the fine-grained individualized representations in the temporal space.
Considering this, we leverage a temporal pyramid pooling (TPP) layer~\cite{wang2016temporal} and our proposed learnable identity weight $W_{id}$ to capture the $IM$ feature.
The TPP layer is used to capture the local representations in the temporal space, and $W_{id}$ is applied to refine the $IM$ information from numerous local representations. The details are shown in Fig.~\ref{fig:dis_generator_block} (a).
In particular, TPP has $B$ scales in the temporal space, where $b \in \mathcal{B}=\{1, 2, 4, \cdots, N\}, len(\mathcal{B})=B$.
The feature extracted by temporal convolution layers divide $B$ group, each group split $2^{b-1}$ bins on the temporal dimension, \textit{i.e.} $\sum_b^{\mathcal{B}}2^{b-1}$ bins in total.
Next, we employ global average pooling (GAP) and global max pooling (GMP) to capture the local representations for each bin.
In other words, the TPP layer leverages multiple temporal pooling layers with different receptive fields to achieve temporal local representations from human movements. 
The TPP layer is followed by identity weight learning. The identity weight is $W_{id} \in \mathbb{R}^{\sum_b^{\mathcal{B}}2^{b-1} \times C_{in} \times C_{out}}$. The design lets the encoder $E_{IM}$ focus on fine-grained $IM$ representations.

The combinations of temporal convolution layers and global average pooling layer have widely been utilized to obtain $HS$ representation~\cite{aberman2019learning, yang2020transmomo, zhu2022mocanet}. 
However, they typically ignore the fact that the $HS$ representation is a spatial-level representation. To address this problem,  we design a spatial-level network to focus on extracting $HS$ representation. 
In particular, $E_{HS}$ has several spatial-level linear layers in place of the traditional temporal convolution layers.

\begin{figure}[t]
    \centering
    \includegraphics[width=1\linewidth]{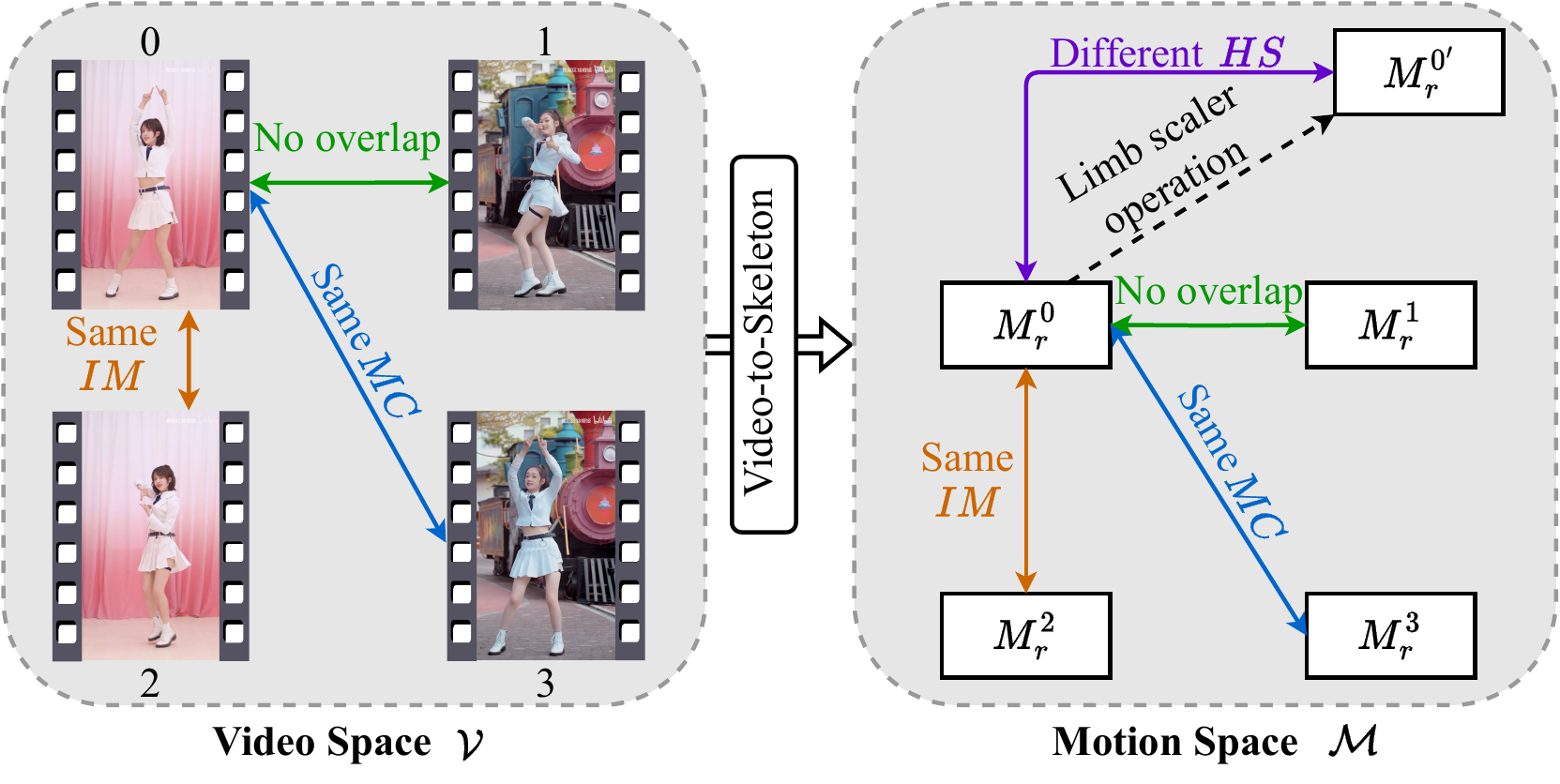}
    \caption{The different data pairs for training. Real datasets typically lack pairs of samples that share the same $HS$ while differing in the other two attributes ($MC$, $IM$). To overcome this limitation, we employ a technique called limb scaler operation~\cite{yang2020transmomo}. This method allows us to artificially create a new sample, denoted as $M_r^{0'}$, which maintains consistency in two attributes ($MC$ and $IM$) but exhibits a different $HS$.}
    \label{fig:data_sampling}
\end{figure}

\subsubsection{Motion generator} \label{sec:generator}

The motion generator is formulated as $M_f = \mathbf{G}(MC, IM, HS)$, which needs three input features achieved by the disentanglement block $\mathbf{E}$. 
Since those three representations have different distributions in the latent space, fusing these representations is one of the major goals for the generator. 
Some previous methods~\cite{aberman2019learning, yang2020transmomo} bypass this issue by concatenating these features directly. 
Those methods only partially explore interactions or correlations among multi-distribution representations but are difficult to provide the generated motions in good quality. 
In our method, the \textit{IBup} block is specifically designed to fuse the multi-distribution representations, and its structure is shown in Fig.~\ref{fig:dis_generator_block} (b). 
In \textit{IBup}, adaptive instance normalization (AdaIN) layers~\cite{adain} are used to fuse the $MC$ and $IM$ representations.
Concretely, the AdaIN layers can achieve mutual integration by normalizing and scaling operations, and it is formulated as follows:
\begin{equation}
    AdaIN(MC, IM) = \sigma(IM) \frac{MC-\mu(MC)}{\sigma(MC)} + \mu(IM),
    \label{adain}
\end{equation} 
where $\sigma$ and $\mu$ are the variance and mean. 

In order to improve the diversity of the synthetic motions, the Gaussian noise is added before AdaIN. 
Moreover, we repeat the $HS$ feature in the temporal dimension and concatenate it in the channel dimension with the aforementioned feature map. 
In order to broadcast $IM$ and $HS$ representations in the generator, we fuse these representations in each scale of the \textit{IBup} block. 
In this manner, it mitigates the issue of $IM$ information from being dissipated after a series of upsampling layers.


\subsection{\textit{IDPres} training schemes}\label{sec:tp}
Inspired by Group Supervised Learning (GSL), we introduce three training schemes: self-reconstruction training, overlap attribute training, and non-overlap attribute training. 
For effective training, it is essential that each subject in the dataset is represented with all attributes necessary for the training process. 
Our collected dataset, $Dancer101$, perfectly aligns with this requirement, encompassing 101 subjects, each performing an identical set of movements. 
Additionally, we have developed a data sampling strategy to support three training schemes.
It is important to note that these phases are executed simultaneously rather than sequentially.

\subsubsection{Data sampling and augmentation} \label{subsubsec:data_sampling}

The videos $\mathcal{V}=\{0, 1, 2, 3\}$ from the dataset are formed into three different pairs for training. The pairs are for the same $MC$, the same $IM$ and non-overlap videos (no attributes overlap), as is shown in Fig.~\ref{fig:data_sampling}.
The 2D skeletons of these videos are extracted by a lightweight pose estimation algorithm, OpenPose~\cite{openpose}, and the extracted motion data is denoted as $\mathcal{M}=\{M_r^0, M_r^1, M_r^2, M_r^3\}$. 
Furthermore, due to the fact that there are no samples in the real dataset with the same $HS$ while differing in the other two attributes ($MC$, $IM$), we employ limb scaler operation~\cite{yang2020transmomo} to augment the samples to meet the requirement. As a result, the real motion space $\mathcal{M}$ can be denoted as $\{M_r^0, M_r^{0'}, M_r^1, M_r^2, M_r^3\}$. 

\begin{figure}[t]
	\centering
    \includegraphics[width=0.8\linewidth]{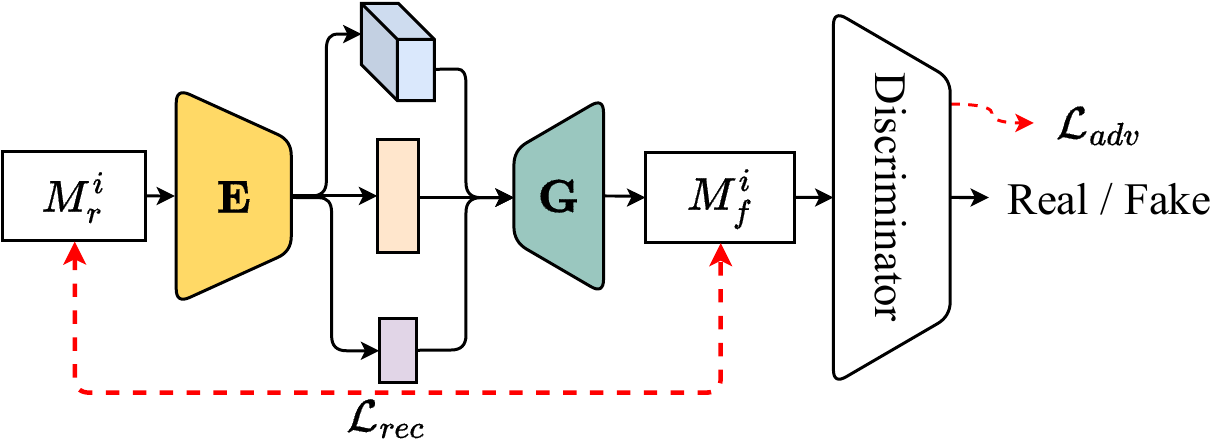}
	\caption{The self-reconstruction pipeline.}
	\label{fig:self-reconstruction}
\end{figure}

\begin{figure*}[t]
\begin{minipage}{0.4\linewidth}
 \centering
 \centerline{\includegraphics[width=0.9\linewidth]{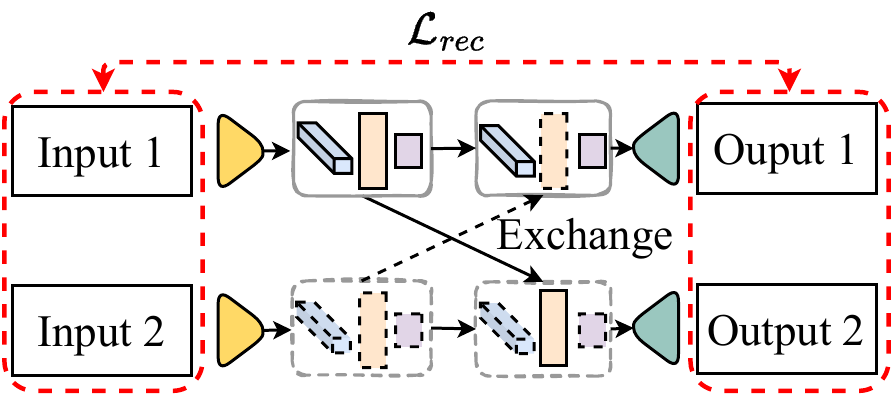}}
 \subcaption{Overlap attribute training pipeline.}
\end{minipage}
\hfill
\begin{minipage}{0.55\linewidth}
\centering
\resizebox{\textwidth}{!}{%
\begin{tabular}{c|c|c|c|c}
\hline
\multirow{2}{*}{Name} & \multirow{2}{*}{\begin{tabular}[c]{@{}c@{}}Exchange \\ content\end{tabular}} & \multirow{2}{*}{(Input 1, Input 2)} & \multirow{2}{*}{(Output 1, Output 2)} & \multirow{2}{*}{$\mathcal{L}_{rec}$}                                             \\
                      &                                                                              &                                     &                                       &                                                                                  \\ \hline
Overlap $MC$          & $MC$                                                                         & $(M_r^0, M_r^3)$                    & $(M_f^{300}, M_f^{033})$              & \begin{tabular}[c]{@{}c@{}}$\mathcal{L}_{rec}^{MC}$ \\ (Eq.~\ref{mc_rec})\end{tabular} \\ 
Overlap $IM$          & $IM$                                                                         & $(M_r^0, M_r^2)$                    & $(M_f^{020}, M_f^{202})$              & \begin{tabular}[c]{@{}c@{}}$\mathcal{L}_{rec}^{IM}$\\  (Eq.~\ref{im_rec})\end{tabular} \\ 
Overlap $HS$          & $HS$                                                                         & $(M_r^0, M_r^{0'})$                   & $(M_f^{000'}, M_f^{0'0'0})$           & \begin{tabular}[c]{@{}c@{}}$\mathcal{L}_{rec}^{HS}$ \\ (Eq.~\ref{body_rec})\end{tabular} \\ \hline
\end{tabular}%
}
\subcaption{Three different cases for inputs, outputs, exchanging contents and loss functions.}
\end{minipage}
\caption{(a) Overlap attribute training pipeline. The yellow trapezoids are the disentanglement encoders, and the green trapezoids are the generators. (b) $M_{f}^{i,j,k}, i,j,k \in \{0, 2, 3, 0'\}$ means generated motions with $MC$ from $M_{r}^{i}$, $IM$ from $M_{r}^{j}$, and $HS$ from $M_{r}^{k}$ respectively. }
\label{fig:overlap_rec}
\end{figure*}

\subsubsection{Self-reconstruction training} \label{subsubsec:self_reconstruction}
To improve the fidelity of motion reconstruction and generation, we introduce a self-reconstruction training pipeline for all sampled motion clips. In this step, the disentanglement block and generator block are used to reconstruct the input motion clips $M_{r}^{i}, i \in\{0, 1, 2, 3, 0'\}$, as shown in Fig.~\ref{fig:self-reconstruction}.
The generated motion clips are termed as $M_{f}^{i}$.
In this training scheme, the ground truth motion clips are $M_{r}^{i}$, since the goal of this stage is to reconstruct each input motion sequence. Thus,
this process is  formulated as:
\begin{equation}
\mathbf{E}(M_f^{i}) = \mathbf{G}(\mathbf{E}(M_r^{i})).
\end{equation}
To further improve the reconstruction quality of our model, 
the reconstruction loss is presented as follows:
\begin{equation}
    \mathcal{L}_{rec}^{i}= \mathbb{E}_{M_{r}^{i} \sim \mathcal{M}}  \| M_{r}^{i} - M_{f}^{i} \|,
    \label{rec_loss}
\end{equation}
where $\mathcal{M}$ represent the space of real motion sequences in Fig.~\ref{fig:dis_generator_block}.
Moreover, to improve the fidelity of synthetic motion clips, a discriminator is used to measure the domain discrepancy between real and synthetic motion clips. 
Therefore, the adversarial loss for $i$-th motion clip is formulated as:
\begin{equation}
    \mathcal{L}_{adv}^{i}= \mathbb{E}_{M_{r}^{i} \sim \mathcal{M}} \left [\frac{1}{2} log D(M_{r}^{i}) + \frac{1}{2} log(1-D(M_{f}^{i})) \right ].
    \label{adv_loss}
\end{equation}

\subsubsection{Overlap attribute training} \label{subsubsec:overlap}
To disentangle each attribute ($MC$, $IM$, or $HS$) in the latent representations efficiently, we swap the attributes in latent space to generate expected motion clips.  Specifically, two motion clips with one attribute with identical values should be reconstructed into nearly the original motion clips when the latent representations for that attribute are swapped,
as shown in Fig.~\ref{fig:overlap_rec}.
For sampling three group motion clips $(M_r^{0}, M_r^{3})$, $(M_r^{0}, M_r^{2})$, $(M_r^{0}, M_r^{0'})$, we conduct the overlap training pipeline and compute reconstruction loss functions of $IM$, $MC$, and $HS$, respectively.
Specifically, regarding the disentanglement of $MC$, when provided with two motion clips (named $M_r^{0}$, and $M_{r}^{3}$) sharing the same dance movements, disentanglement block (Fig.~\ref{fig:dis_generator_block} (a), we named it as $\mathbf{E}$), used to respectively obtain two motion clips' $MC$, $IM$, and $HS$ representations, named $(MC^0, IM^0, HS^0)$ and $(MC^3, IM^3, HS^3)$.
This process  is  formulated as:
\begin{equation}
\begin{aligned}
(MC^0, IM^0, HS^0) &= \mathbf{E}(M_r^{0}), \\
(MC^3, IM^3, HS^3) &= \mathbf{E}(M_r^{3}).
\end{aligned}
\end{equation}
Then, the $MC^0$ and $MC^3$ representations are swapped in latent space, \textit{i.e.}  $(MC^3, IM^0, HS^0)$ and $(MC^0, IM^3, HS^3)$ are obtained.
Next, the generator block (Fig.~\ref{fig:dis_generator_block} (b), we named it as $\mathbf{G}$) is employed to generate motion clips that have corresponding attributes for the two groups' representation.
This process can be formulated as:
\begin{equation}
\begin{aligned}
M_f^{300} &= \mathbf{G}((MC^3, IM^0, HS^0)), \\
M_f^{033} &= \mathbf{G}((MC^0, IM^3, HS^3)).
\end{aligned}
\end{equation}
Similarly, the $IM$ and $HS$ undergo the same process with $MC$.
The overlap-$IM$ training schemes can be formulated as:
\begin{equation}
\begin{aligned}
&(MC^0, IM^0, HS^0) = \mathbf{E}(M_r^{0}), \\
&(MC^2, IM^2, HS^2) = \mathbf{E}(M_r^{2}), \\
&M_f^{020} = \mathbf{G}((MC^0, IM^2, HS^0)), \\
&M_f^{202} = \mathbf{G}((MC^2, IM^0, HS^2)).
\end{aligned}
\end{equation}
The overlap-$HS$ training schemes can be formulated as:
\begin{equation}
\begin{aligned}
&(MC^0, IM^0, HS^0) = \mathbf{E}(M_r^{0}), \\
&(MC^{0'}, IM^{0'}, HS^{0'}) = \mathbf{E}(M_r^{0'}), \\
&M_f^{000'} = \mathbf{G}((MC^0, IM^0, HS^{0'})), \\
&M_f^{0'0'0} = \mathbf{G}((MC^{0'}, IM^{0'}, HS^0)).
\end{aligned}
\end{equation}
Then, three motion consistency loss functions are employed: the $MC$ consistency reconstruction loss, the $IM$ consistency reconstruction loss, and the $HS$ consistency reconstruction loss.
Their definitions are as follows:
\begin{equation}
    \mathcal{L}_{rec}^{MC}= \| M_{r}^{0} - M_{f}^{300} \|+\| M_{r}^{3} - M_{f}^{033} \|,
    \label{mc_rec}
\end{equation} 
\begin{equation}
    \mathcal{L}_{rec}^{IM}= \| M_{r}^{0} - M_{f}^{020} \|+\| M_{r}^{2} - M_{f}^{202} \|,
    \label{im_rec}
\end{equation} 
\begin{equation}
    \mathcal{L}_{rec}^{HS}= \| M_{r}^{0'} - M_{f}^{000'} \|+\| M_{r}^{0} - M_{f}^{0'0'0} \|,
    \label{body_rec}
\end{equation} 
where, $M_{f}^{i,j,k}, i,j,k \in \{0, 2, 3, 0'\}$ means generated motion clips with $MC$ from $M_{r}^{i}$, $IM$ from $M_{r}^{j}$, and $HS$ from $M_{r}^{k}$ respectively.

In addition, in the field of disentanglement~\cite{hou2021learning,gal2021mrgan,wang2022disentangled}, the metric loss is also widely used. 
To obtain expected disentanglement performance, the triplet loss~\cite{hermans2017defense} is used in our disentanglement method. 
Specifically, a triplet is denoted as $m=(\tau, \upsilon, \psi)$, where $\tau$ represents the anchor feature, $\upsilon$ is a feature with the same label as the anchor $\tau$ and $\psi$ is a feature with a label different from that of the anchor $\tau$. 
Thus, the triplet loss is formulated as follows:
\begin{equation}
    \mathcal{L}_{tri}(m)= max\{0, \delta + D_{\tau, \upsilon}-D_{\tau, \psi}\},
    \label{ba+}
\end{equation}
where $\delta$ is the margin parameter, $D_{\tau, \upsilon}$ and $D_{\tau, \psi}$ are the intra-class distance and the inter-class distance respectively. 
Note that three triplet losses are applied to decompose three different latent representations, and the details of the training algorithm are shown in Algorithm~\ref{train_regime}.

\begin{figure}[t]
	\centering
       	\includegraphics[width=0.8\linewidth]{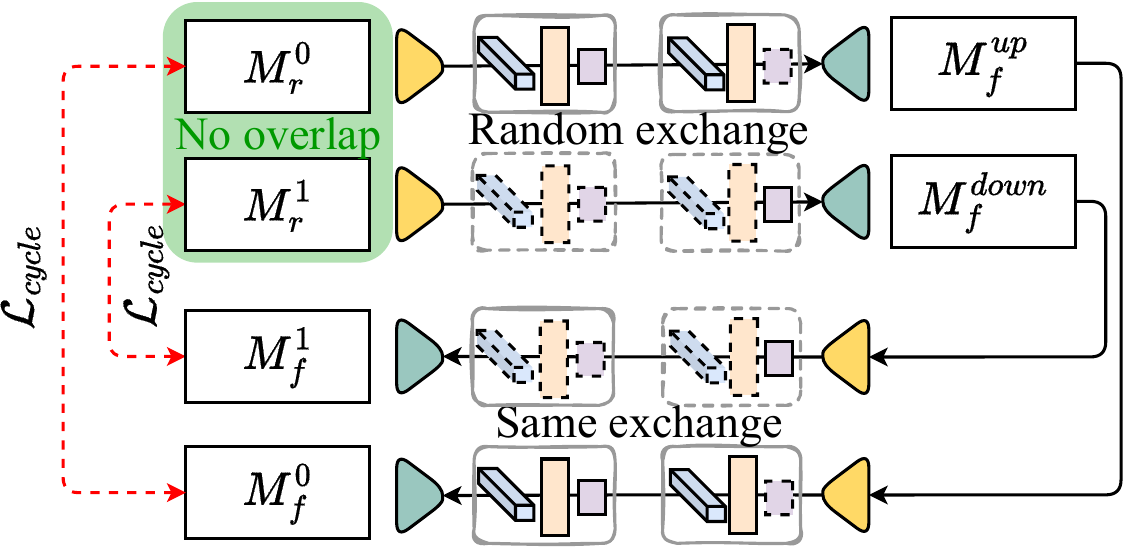}
	\caption{The overview of non-overlap attribute training pipeline. Given any examples without overlapping attributes, we randomly exchange an attribute.}
	\label{fig:non-overlap}
\end{figure}

\begin{algorithm} [t]
	\caption{Training regime.}
	\label{train_regime}
 \footnotesize
	\hspace*{0.1in} {\bf Input:} 
    Real motion dataset $\mathcal{M}$\\
    \hspace*{0.02in} {\bf Output:} 
    Loss items: $\mathcal{L}_{cycle}$, $\mathcal{L}_{rec}^{IM}$, $\mathcal{L}_{rec}^{MC}$, $\mathcal{L}_{rec}^{HS}$, $\mathcal{L}_{rec}$, $\mathcal{L}_{tri}$, $\mathcal{L}_{adv}$
	\begin{algorithmic}[1]
        \STATE Sample motions from $\mathcal{M}$, which are regarded as $M_{r}^{i}, i \in \{0,1,2,3\}$
        \STATE For $M_{r}^{0}$, using the limb scaler operation and obtaining the motion regarded as $M_{r}^{0'}$
        \STATE \textcolor[rgb]{0.1,0.4,0.3}{\# Self-reconstruction training}
        \FOR {$i \in \{0,1,2,3,0'\}$}
            \STATE $MC^i \leftarrow E_{MC}(M_{r}^{i})$
            \STATE $IM^i \leftarrow E_{IM}(M_{r}^{i})$
            \STATE $HS^i \leftarrow E_{HS}(M_{r}^{i})$
            \STATE $M_{f}^{i} \leftarrow G(MC^i, IM^i, HS^i)$
            \STATE $\mathcal{L}_{rec} \leftarrow \mathcal{L}_{rec} + \mathbb{E}\left [\| M_{r}^{i} - M_{f}^{i} \| \right ]$
            \STATE $\mathcal{L}_{adv} \leftarrow \mathcal{L}_{adv} + \mathbb{E}\left [\frac{1}{2} log D(M_{r}^{i}) + \frac{1}{2} log(1-D(M_{f}^{i})) \right ]$
        \ENDFOR
        \\
        \textcolor[rgb]{0.5,0.8,0.7}{\# Computing triplet loss by Equation~\ref{ba+}}
        \STATE $\mathcal{L}_{tri} \leftarrow \mathcal{L}_{tri}((MC^0, MC^3, MC^2)) + $
        \\
        \qquad \qquad $\mathcal{L}_{tri}((IM^0, IM^2, IM^3)) + $ 
        \\
        \qquad \qquad $\mathcal{L}_{tri}((HS^0, HS^2, HS^{0'}))$
        \STATE \textcolor[rgb]{0.1,0.4,0.3}{\# Overlap attribute training}
        \\
        \textcolor[rgb]{0.5,0.8,0.7}{\# Overlap $MC$}
        \STATE $M_{f}^{300} \leftarrow \mathbf{G}(MC^3, IM^0, HS^0)$
        \STATE $M_{f}^{033} \leftarrow \mathbf{G}(MC^0, IM^3, HS^3)$
        \STATE $\mathcal{L}_{rec}^{MC} \leftarrow$ Equation~\ref{mc_rec} $\leftarrow (M_{f}^{300}, M_{f}^{033}, M_{r}^{0}, M_{r}^{3})$
        \\
        \textcolor[rgb]{0.5,0.8,0.7}{\# Overlap $IM$}
        \STATE $M_{f}^{020} \leftarrow \mathbf{G}(MC^0, IM^2, HS^0)$
        \STATE $M_{f}^{202} \leftarrow \mathbf{G}(MC^2, IM^0, HS^2)$
        \STATE $\mathcal{L}_{rec}^{IM} \leftarrow$ Equation~\ref{im_rec} $\leftarrow (M_{f}^{020}, M_{f}^{202}, M_{r}^{0}, M_{r}^{2})$
        \\
        \textcolor[rgb]{0.5,0.8,0.7}{\# Overlap $HS$}
        \STATE $M_{f}^{000'} \leftarrow \mathbf{G}(MC^0, IM^0, HS^{0'})$
        \STATE $M_{f}^{0'0'0} \leftarrow \mathbf{G}(MC^{0'}, IM^{0'}, HS^0)$
        \STATE $\mathcal{L}_{rec}^{HS} \leftarrow$ Equation~\ref{body_rec} $\leftarrow (M_{f}^{000'}, M_{f}^{0'0'0}, M_{r}^{0}, M_{r}^{0'})$
        \STATE \textcolor[rgb]{0.1,0.4,0.3}{\# Non-overlap attribute training}
        
        \STATE $\beta \sim \{IM, MC, HS\}$ \textcolor[rgb]{0.5,0.8,0.7} {\# Random sample an attribute}
        \STATE $z^0 \leftarrow (MC^0, IM^0, HS^0)$, 
        \\
        $z^1 \leftarrow (MC^1, IM^1, HS^1)$
        \STATE $(M_{f}^{up}, M_{f}^{down}) \leftarrow \mathbf{G}(\mathrm{Swap}(z^0, z^1, \beta))$
        \STATE $\hat{z}^0 \leftarrow (E_{MC}(M_{f}^{up}), E_{IM}(M_{f}^{up}), E_{HS}(M_{f}^{up}))$, 
        \\
        $\hat{z}^1 \leftarrow (E_{MC}(M_{f}^{down}), E_{IM}(M_{f}^{down}), E_{HS}(M_{f}^{down}))$
        \STATE $(M_{f}^{0}, M_{f}^{1}) \leftarrow \mathbf{G}(\mathrm{Swap}(\hat{z}^0, \hat{z}^1, \beta))$
        \STATE $\mathcal{L}_{cycle} \leftarrow$ Equation~\ref{cycle_loss} $\leftarrow (M_{f}^{0}, M_{f}^{1}, M_{r}^{0}, M_{r}^{1})$
	\end{algorithmic}
\end{algorithm}

\subsubsection{Non-overlap attribute training} \label{subsubsec:nonoverlap}

To handle the case of no attribute paired, we introduce a cycle-based reconstruction training pipeline. 
This pipeline is implemented on all example pairs, regardless of whether they share an attribute or not. 
However, we may not have ground truth for generated motion clips in our training set.
For example, we do not have samples of both $M_{r}^{1}$'s $HS$ and $M_{r}^{0}$'s $IM$ in the dataset when swapping the $HS$ attribute. 
To address the problem, inspired by CycleGAN~\cite{zhu2017unpaired}, we re-decompose and re-generate for generated motion clips, as is shown in Fig.~\ref{fig:non-overlap}. 
Given a paired of no attribute overlapped motion clips $M_{r}^{0}$ and $M_{r}^{1}$, we first obtain their corresponding representations by the disentanglement block, then randomly exchange an attribute's representation ($IM$, $MC$, or $HS$), finally generate motion clips by generator. 
In order to recover the generated motion clips, we employ the same exchanging operation, \textit{i.e.} if the $MC$ attribute is exchanged in the first exchange stage, the $MC$ attribute is also exchanged in the second stage.
The formula of this process is shown in Algorithm~\ref{train_regime}, lines 24 to 28.
Finally, we use the cycle reconstruction loss to optimize the parameters of the networks, which are formulated as follows:
\begin{equation}
    \mathcal{L}_{cycle}= \mathbb{E}_{i \sim \{0,1\}} \| M_{r}^{i} - M_{f}^{i} \|.
    \label{cycle_loss}
\end{equation}

\begin{figure}[t]
	\centering
	\includegraphics[width=\linewidth]{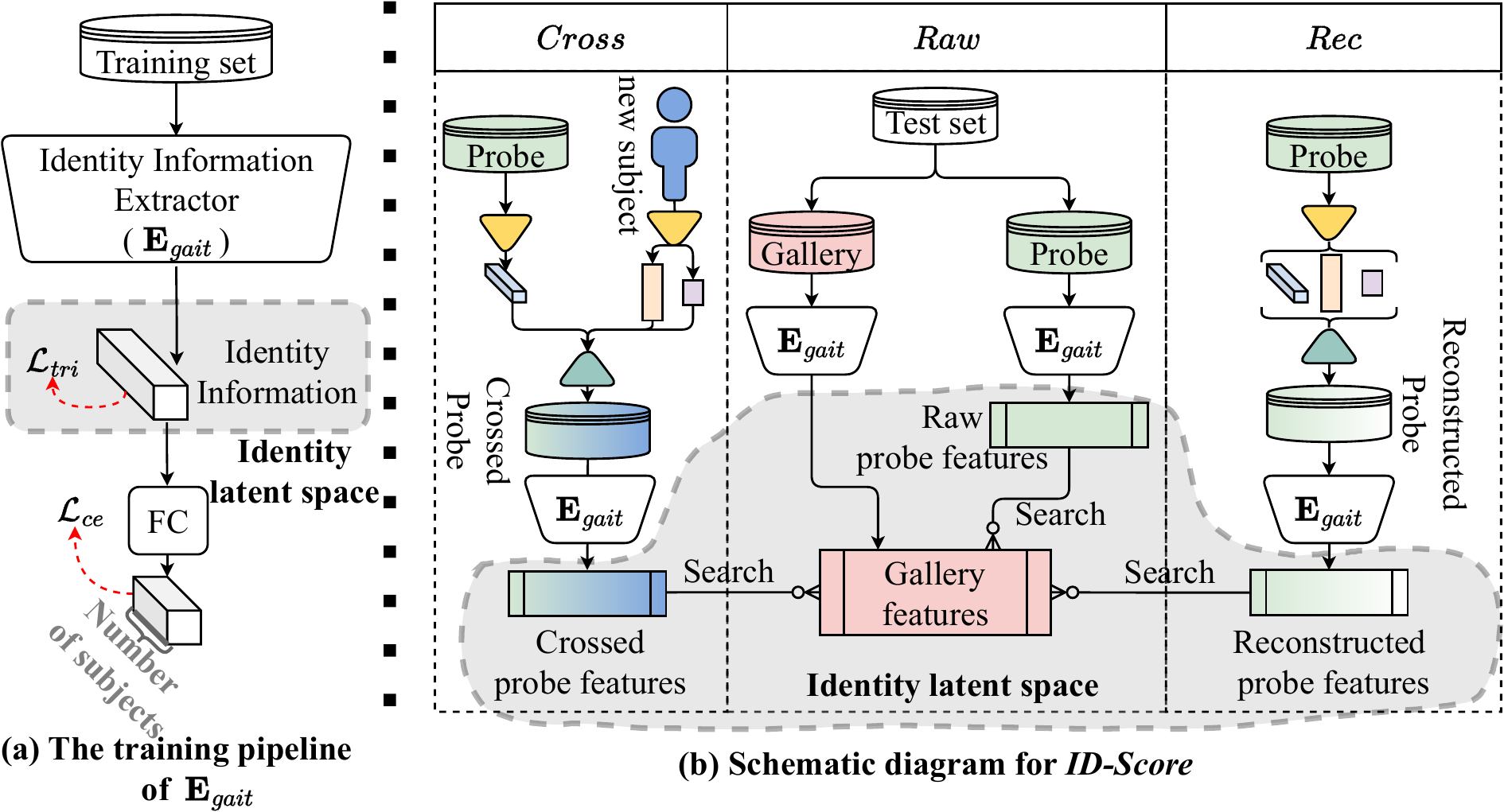}
	\caption{The overview of the metric of \textit{ID-Score}. (a) The training pipeline of $\mathbf{E}_{gait}$. $\mathcal{L}_{tri}$ means triplet loss function, and $\mathcal{L}_{ce}$ means cross-entropy loss function. FC means the fully connected layers. Typically, in gait recognition methods, they just use identity information (the representation before the FC layer) to conduct evaluation and recognition, the details can be found in \cite{fan2023opengait}. (b) The schematic diagram for \textit{ID-Score}. The $Raw$, $Cross$, and $Rec$ denote the raw data stage, cross-stage, and reconstruction stage evaluation pipeline, respectively.}
    \label{fig:id_score}
\end{figure}

It is important to note that these phases are executed simultaneously rather than sequentially. 
The disentanglement block and the motion generator in all training schemes share the parameters. 
The complete training regime is shown in Algorithm~\ref{train_regime}.

\begin{table}[t]
\centering
\caption{The recognition accuracy of three gait recognition methods on different datasets.}
\label{tab:gait_ex}
\resizebox{0.45\textwidth}{!}{%
\begin{tabular}{c|ccc}
\hline
\diagbox{Dataset}{Gait methods}           & GaitGraph~\cite{teepe2021gaitgraph} & GaitGraph 2~\cite{teepe2022gaitgraph2} & Gait-TR~\cite{zhang2023spatial} \\ \hline\hline
$Dancer101$ & 60.00\%   & 73.06\%     & 81.22\% \\
Mixamo~\cite{mixamo}      & 93.59\%   & 96.79\%     & 98.72\% \\
UTD-MHAD~\cite{chen2015utd}    & 97.06\%   & 97.06\%     & 100\%   \\ \hline
\end{tabular}%
}
\end{table}

\subsection{\textit{ID-Score} metric} \label{sec:idscore}

Based on previous training, \textit{IDPres} can decompose $IM$, $MC$, and $HS$ from dancing movements, and recompose them to generate new dancing movements. 
To quantify the preservation of $IM$ information, we have developed a novel evaluation metric known as the Identity-Score (\textit{ID-Score}), which is based on gait recognition.
Gait recognition has successfully demonstrated that leveraging subjects’ identity information (both $IM$ and $HS$) in walking is beneficial to identifying individuals.
We assume that gait recognition methods can similarly represent identity information from dance movements. 
To validate this point, we train all three gait recognition methods on three different dance datasets via pipeline in Fig.~\ref{fig:id_score} (a), with results detailed in Table~\ref{tab:gait_ex}.
Notably, the recognition accuracy was 81.22\% for Dancer101, 98.72\% for Mixamo~\cite{mixamo}, and 100.00\% for UTD-MHAD~\cite{chen2015utd} using the skeleton-based gait recognition model Gait-TR~\cite{zhang2023spatial}. 
These high recognition accuracies indicate that gait recognition methods are indeed effective in extracting identity information information from dance movements.

The \textit{ID-Score} metric operates through three stages: raw data evaluation ($Raw$), reconstruction data evaluation ($Rec$), and cross-subject identity information evaluation ($Cross$), as illustrated in Fig.~\ref{fig:id_score}. 
In the $Raw$ stage, an identity information extractor ($\mathbf{E}_{gait}$) is used to map both the raw probe data and gallery data into an identity latent space. 
The similarity between each motion clip in the probe set and the gallery set is computed to predict identity labels.
The recognition accuracy, known as $Rank_1$, evaluates this stage. 
During the $Rec$ stage, probe data is reconstructed using the evaluated HMT methods, and then projected into the identity latent space using the $\mathbf{E}_{gait}$. 
This stage's accuracy, $Rank_1^{rec}$, is then calculated. 
In the $Cross$ stage, a new subject performs as the target subject, and all probe data as source subjects, leveraging \textit{IDPres} to transfer the target's $IM$ and $HS$ information to each source.
Then, the generated motion clips are projected into the identity latent space via $\mathbf{E}_{gait}$.
The resulting accuracy, $Rank_1^{cross}$, is expected to be lower since the new subject is not registered in the gallery set. 
Moreover, the crucial point is that poor reconstruction performance can also diminish $Rank_1^{cross}$ performance. 
Therefore, the difference between $Rank_1^{rec}$ and $Rank_1^{cross}$ is used to counteract this issue, resulting in the \textit{ID-Score}. 
Higher \textit{ID-Score} values indicate better preservation of $IM$ information information. 
\begin{equation}
   \textit{ID-Score} = Rank_1^{rec} - Rank_1^{cross}.
\end{equation}

\section{Experiments}

\subsection{Datasets} \label{dancer101}

In this paper, we utilize four different datasets, Mixamo~\cite{mixamo}, UTD-MHAD~\cite{chen2015utd}, our dataset $Dancer101$ and $Dancer99$, to train the models and evaluate the performance of them. $Dancer99$ is used to evaluate the cross-domain ability for different methods. $Dancer101$ and $Dancer99$ are described in Section~\ref{sec:data_pre}.

Adobe's Mixamo~\cite{mixamo} provides animated 3D characters for games, films, and other applications. The data is fully virtual data and is not collected from a real scenario. Inspired by~\cite{aberman2019learning}, we project the animated 3D models into 2D from 7 different view angles. 
We divide each motion into several motion animations of 64 frames. 
Then, we also split the datasets into a training set and a test set. 
The training set contains 32 subjects, and each subject contains 799 motion animations. 
The testing set contains 4 subjects, and each subject contains 64 motion animations. 

UTD-MHAD dataset~\cite{chen2015utd} was collected in a real scenario for action recognition. So, we used a similar processing with $Dancer101$ on UTD-MHAD. 
The UTD-MHAD dataset has 8 subjects, and each subject has 27 motion animations. 
All the subjects have been split into 6 for training and 2 for testing without overlap. 
The statistics of the 3 datasets are shown in Table~\ref{tab:dataset}.


\subsection{Experimental Settings}

\subsubsection{Implementation details} 
In the training stage, the number of input frames is $T = 64$, and the number of joints is 15. 
In the inference phase, our proposed method can handle videos of variable lengths.
In the disentanglement block, we set the numbers of channels of the outputs as $C_{MC}=128, C_{IM}=128, C_{HS}=8$ respectively.
The setting of TPP scalers is the same to~\cite{chao2021gaitset}, \textit{i.e.} $\mathcal{B}=\{1, 2, 4, 8, 16\}$. 
In the discriminator, four 1D temporal convolution layers are employed, and the number of the last channel of the discriminator is $128$.
The adversarial losses are adopted from least squares generative adversarial networks (LSGAN)~\cite{mao2017least}, and they can constrain the distribution of generated motions similarly to real motions.

In the skeleton-to-video render stage, we adopt the VQGAN~\cite{vq-gan} model as the backbone, which is originally for image-to-image translation, to render skeletons to RGB images frame by frame. 
In order to handle the one-by-one translation problem~\cite{wang2019few}, \textit{i.e.} a subject corresponds to a model parameter, additional conditional images are introduced.
Those conditional images mainly provide the RGB information to the model.
In particular, we randomly choose an image with the same subject as the conditional input, and that concatenates the channel axes of the conditional and skeleton images as input to the VQGAN model.
The resolution of all images is $512 \times 512$ in our experiments, and other settings for hyperparameters are strictly following~\cite{vq-gan}.

We utilize a model-based gait recognition method, GaitGraph~\cite{teepe2021gaitgraph}, GaitGraph 2~\cite{teepe2022gaitgraph2}, and Gait-TR~\cite{zhang2023spatial}, to evaluate the synthetic motion clips.
The gait recognition models are trained on the training sets of different datasets. 

\subsubsection{Evaluation metrics}
Four quantitative metrics are involved to evaluate the performance of our method and SOTA methods.
The metrics are mean square error (MSE), mean average error (MAE), and Frechet Motion Distance(FMD)~\cite{heusel2017gans}, and the proposed \textit{ID-Score}.
For the FMD metric, we employ GaitGraph~\cite{teepe2021gaitgraph} to achieve the inception features, which can be formulated as follows:
\begin{equation}
    \begin{split}
        FMD &= \| \mu_{\mathcal{F}(M_r)} - \mu_{\mathcal{F}(M_f)} \| ^2 \\
        &- Tr(\textstyle\sum_{\mathcal{F}(M_r)} + \textstyle \sum_{\mathcal{F}(M_f)} - 2 \textstyle\sum_{\mathcal{F}(M_r)} \textstyle\sum_{\mathcal{F}(M_f)}),
    \end{split}
    \label{fmd}
\end{equation}
where, $\mu_{\mathcal{F}(M_r)}$ represents the mean and standard deviation of $\mathcal{F}(M_r)$, $\mathcal{F}(M_r)$ means the GaitGraph feature of real motion, $Tr$ is the trace of the matrix, and $\sum_{\mathcal{F}(M_r)}$ is the covariance matrix of the feature map.

\subsection{Comparisons with state-of-the-art methods}
We compared the proposed \textit{IDPres} with TransMoMo~\cite{yang2020transmomo}, Learning Character-Agnostic Motion (LCM)~\cite{aberman2019learning}, MoCaNet~\cite{zhu2022mocanet} on several aspects. 

\subsubsection{Evaluation of reconstruction quality}

\begin{table}[t]
\centering
\caption{The reconstruction metrics on the three datasets.}
\label{tab:reconstruction-tab}
\resizebox{0.45\textwidth}{!}{%
\begin{tabular}{c|c|c|c|c}
\hline
Datasets                     & Methods   & MSE($\downarrow$) & MAE($\downarrow$) & FMD($\downarrow$) \\ \hline\hline
\multirow{4}{*}{Mixamo}      & TransMoMo & 0.023             & 0.115             & 0.343             \\ \cline{2-5} 
                             & LCM       & 0.354             & 0.392             & 0.269             \\ \cline{2-5} 
                             & MocaNet   & \textbf{0.007}    & \textbf{0.054}    & 0.017             \\ \cline{2-5} 
                             & Ours      & 0.023             & 0.097             & \textbf{0.005}    \\ \hline\hline
\multirow{4}{*}{UTD-MHAD}    & TransMoMo & 0.342             & 0.433             & 0.037             \\ \cline{2-5} 
                             & LCM       & 0.550             & 0.520             & 0.104             \\ \cline{2-5} 
                             & MocaNet   & 0.314             & \textbf{0.285}    & 0.024             \\ \cline{2-5} 
                             & Ours      & \textbf{0.284}    & 0.340             & \textbf{0.014}    \\ \hline\hline
\multirow{4}{*}{$Dancer101$} & TransMoMo & 0.184             & 0.300             & 0.129             \\ \cline{2-5} 
                             & LCM       & 0.514             & 0.511             & 0.146             \\ \cline{2-5} 
                             & MocaNet   & 0.188             & \textbf{0.244}    & 0.081             \\ \cline{2-5} 
                             & Ours      & \textbf{0.152}    & 0.267             & \textbf{0.055}    \\ \hline
\end{tabular}%
}
\end{table}

In order to quantitatively compare the motion reconstruction quality of \textit{IDPres} with that of the previous methods, three well-accepted metrics are adopted: mean square error (MSE), mean average error (MAE), and Frechet Motion Distance (FMD).  Table~\ref{tab:reconstruction-tab} offers motion reconstruction quantitative results, our method generally performs better than previous methods. We see that MoCaNet outperforms our method in the MAE, but our method reduces FMD values when compared to MoCaNet. Note that FMD is a significant measure to evaluate the identity preservation of reconstructed motion clips. The results demonstrate the effectiveness of the method in motion reconstruction.

\begin{table*}[t]
\centering
\caption{Experimental results for \textit{ID-Score} metrics. While Raw represents the upper bound of performance for the current dataset.}
\label{tab:idscore-tab}
\resizebox{\textwidth}{!}{%
\begin{tabular}{c|c||ccc||ccc||ccc}
\hline
\multirow{3}{*}{Datasets}    & Gait models                                                                   & \multicolumn{3}{c||}{GaitGraph~\cite{teepe2021gaitgraph}}                                                             & \multicolumn{3}{c||}{GaitGraph 2~\cite{teepe2022gaitgraph2}}                                                           & \multicolumn{3}{c}{Gait-TR~\cite{zhang2023spatial}}                                                                \\ \cline{2-11} 
                             & \multirow{2}{*}{\begin{tabular}[c]{@{}c@{}}Generative\\ methods\end{tabular}} & \multicolumn{1}{c|}{$Rank_1^{rec}$}   & \multicolumn{1}{c|}{$Rank_1^{cross}$}   & \textit{ID-Score}       & \multicolumn{1}{c|}{$Rank_1^{rec}$}   & \multicolumn{1}{c|}{$Rank_1^{cross}$}   & \textit{ID-Score}       & \multicolumn{1}{c|}{$Rank_1^{rec}$}   & \multicolumn{1}{c|}{$Rank_1^{cross}$}   & \textit{ID-Score}       \\
                             &                                                                               & \multicolumn{1}{c|}{(\%)($\uparrow$)} & \multicolumn{1}{c|}{(\%)($\downarrow$)} & (\%)($\uparrow$) & \multicolumn{1}{c|}{(\%)($\uparrow$)} & \multicolumn{1}{c|}{(\%)($\downarrow$)} & (\%)($\uparrow$) & \multicolumn{1}{c|}{(\%)($\uparrow$)} & \multicolumn{1}{c|}{(\%)($\downarrow$)} & (\%)($\uparrow$) \\ \hline\hline
\multirow{5}{*}{Mixamo}      & TransMoMo                                                                     & \multicolumn{1}{c|}{67.31}            & \multicolumn{1}{c|}{\textbf{25.00}}     & 42.31            & \multicolumn{1}{c|}{83.97}            & \multicolumn{1}{c|}{\textbf{21.79}}     & \textbf{62.18}   & \multicolumn{1}{c|}{76.92}            & \multicolumn{1}{c|}{25.00}              & 51.92            \\ \cline{2-11} 
                             & LCM                                                                           & \multicolumn{1}{c|}{57.69}            & \multicolumn{1}{c|}{32.69}              & 25.00            & \multicolumn{1}{c|}{58.97}            & \multicolumn{1}{c|}{39.74}              & 19.23            & \multicolumn{1}{c|}{73.72}            & \multicolumn{1}{c|}{46.15}              & 27.57            \\ \cline{2-11} 
                             & MoCaNet                                                                       & \multicolumn{1}{c|}{76.28}            & \multicolumn{1}{c|}{27.56}              & 48.72            & \multicolumn{1}{c|}{83.97}            & \multicolumn{1}{c|}{23.08}              & 60.89            & \multicolumn{1}{c|}{96.15}            & \multicolumn{1}{c|}{25.00}              & 71.15            \\ \cline{2-11} 
                             & Ours                                                                          & \multicolumn{1}{c|}{\textbf{91.03}}   & \multicolumn{1}{c|}{25.64}              & \textbf{65.39}   & \multicolumn{1}{c|}{\textbf{87.18}}   & \multicolumn{1}{c|}{25.00}              & \textbf{62.18}   & \multicolumn{1}{c|}{\textbf{98.08}}   & \multicolumn{1}{c|}{\textbf{25.00}}      & \textbf{73.08}   \\ \cline{2-11} 
                             & Raw                                                                           & \multicolumn{3}{c||}{Rank 1: 93.59\%}                                                               & \multicolumn{3}{c||}{Rank 1: 96.79\%}                                                               & \multicolumn{3}{c}{Rank 1: 98.72\%}                                                                \\ \hline\hline
\multirow{5}{*}{UTD-MHAD}    & TransMoMo                                                                     & \multicolumn{1}{c|}{79.41}            & \multicolumn{1}{c|}{52.94}              & 26.47            & \multicolumn{1}{c|}{79.55}            & \multicolumn{1}{c|}{50.00}              & 29.55            & \multicolumn{1}{c|}{84.09}            & \multicolumn{1}{c|}{50.00}              & 34.09            \\ \cline{2-11} 
                             & LCM                                                                           & \multicolumn{1}{c|}{82.35}            & \multicolumn{1}{c|}{64.71}              & 17.64            & \multicolumn{1}{c|}{76.47}            & \multicolumn{1}{c|}{55.88}              & 20.59            & \multicolumn{1}{c|}{77.27}            & \multicolumn{1}{c|}{61.36}              & 15.91            \\ \cline{2-11} 
                             & MoCaNet                                                                       & \multicolumn{1}{c|}{89.71}            & \multicolumn{1}{c|}{52.94}              & 36.77            & \multicolumn{1}{c|}{85.29}            & \multicolumn{1}{c|}{47.06}              & 38.23            & \multicolumn{1}{c|}{84.09}            & \multicolumn{1}{c|}{50.00}              & 34.09            \\ \cline{2-11} 
                             & Ours                                                                          & \multicolumn{1}{c|}{\textbf{91.18}}   & \multicolumn{1}{c|}{\textbf{50.00}}     & \textbf{41.18}   & \multicolumn{1}{c|}{\textbf{94.12}}   & \multicolumn{1}{c|}{\textbf{50.00}}      & \textbf{44.12}   & \multicolumn{1}{c|}{\textbf{95.45}}   & \multicolumn{1}{c|}{\textbf{50.00}}      & \textbf{45.45}   \\ \cline{2-11} 
                             & Raw                                                                           & \multicolumn{3}{c||}{Rank 1: 97.06\%}                                                               & \multicolumn{3}{c||}{Rank 1: 97.06\%}                                                               & \multicolumn{3}{c}{Rank 1: 100.0\%}                                                                \\ \hline\hline
\multirow{5}{*}{$Dancer101$} & TransMoMo                                                                     & \multicolumn{1}{c|}{36.33}            & \multicolumn{1}{c|}{10.41}              & 25.92            & \multicolumn{1}{c|}{32.86}            & \multicolumn{1}{c|}{17.55}              & 15.31            & \multicolumn{1}{c|}{34.69}            & \multicolumn{1}{c|}{18.37}              & 16.32            \\ \cline{2-11} 
                             & LCM                                                                           & \multicolumn{1}{c|}{30.61}            & \multicolumn{1}{c|}{18.57}              & 12.04            & \multicolumn{1}{c|}{30.20}            & \multicolumn{1}{c|}{23.47}              & 6.73             & \multicolumn{1}{c|}{34.08}            & \multicolumn{1}{c|}{22.65}              & 11.43            \\ \cline{2-11} 
                             & MoCaNet                                                                       & \multicolumn{1}{c|}{34.90}            & \multicolumn{1}{c|}{\textbf{10.00}}     & 24.90            & \multicolumn{1}{c|}{41.63}            & \multicolumn{1}{c|}{15.51}              & 26.12            & \multicolumn{1}{c|}{47.96}            & \multicolumn{1}{c|}{16.53}              & 31.43            \\ \cline{2-11} 
                             & Ours                                                                          & \multicolumn{1}{c|}{\textbf{46.53}}   & \multicolumn{1}{c|}{10.61}              & \textbf{35.92}   & \multicolumn{1}{c|}{\textbf{55.92}}   & \multicolumn{1}{c|}{\textbf{10.00}}      & \textbf{45.92}   & \multicolumn{1}{c|}{\textbf{53.06}}   & \multicolumn{1}{c|}{\textbf{10.20}}      & \textbf{42.86}   \\ \cline{2-11} 
                             & Raw                                                                           & \multicolumn{3}{c||}{Rank 1: 60.00\%}                                                               & \multicolumn{3}{c||}{Rank 1: 73.06\%}                                                               & \multicolumn{3}{c}{Rank 1: 81.22\%}                                                                \\ \hline
\end{tabular}%
}
\end{table*}

\begin{figure}[t]
	\centering
	    \includegraphics[width=0.9\linewidth]{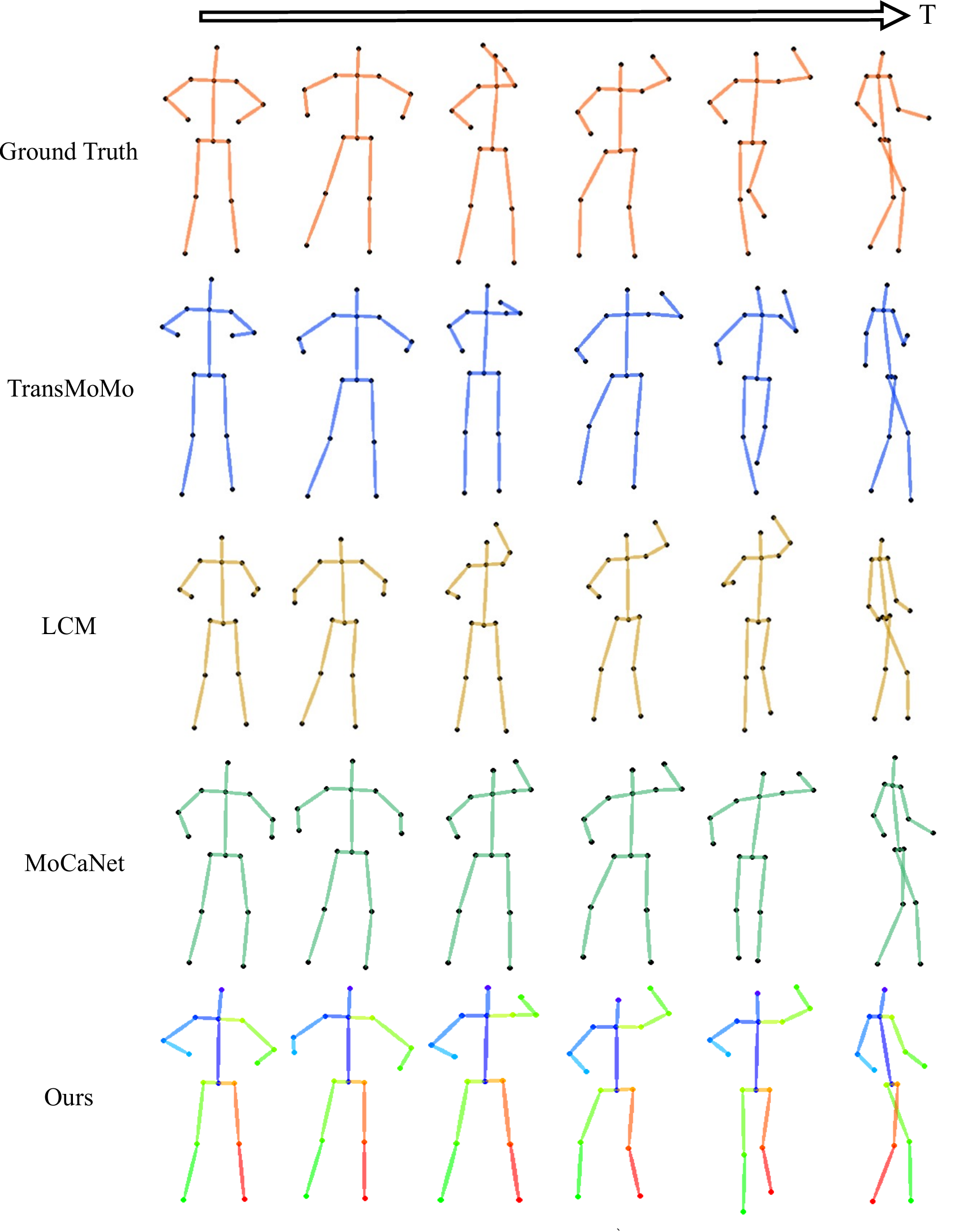}
	\caption{The reconstructed skeletons by different methods. The proposed \textit{IDPres} provides the best reconstruction quality.}
	\label{fig:recon_vis}
\end{figure}

In order to analyze the reconstruction quality visually, Fig.~\ref{fig:recon_vis} provides representative motion reconstruction examples of our \textit{IDPres} and other three comparable methods, where examples are selected from the $Dancer101$ dataset. We observe that the reconstructed skeletons by \textit{IDPres} are more similar to the ground truth than those by previous methods.

\begin{table}[t]
\centering
\caption{The results of cross-domain on the in-the-wild dataset $Dancer99$.}
\label{tab:in-the-wild}
\resizebox{0.45\textwidth}{!}{%
\begin{tabular}{c|c|ccc}
\hline
\multirow{2}{*}{Methods} & $Rank_1^{rec}$   & \multicolumn{1}{c|}{\multirow{2}{*}{MSE($\downarrow$)}} & \multicolumn{1}{c|}{\multirow{2}{*}{MAE($\downarrow$)}} & \multirow{2}{*}{FMD($\downarrow$)} \\
                         & (\%)($\uparrow$) & \multicolumn{1}{c|}{}                                   & \multicolumn{1}{c|}{}                                   &                                    \\ \hline\hline
TransMoMo                & 6.06             & \multicolumn{1}{c|}{0.17}                               & \multicolumn{1}{c|}{0.28}                               & 0.20                               \\ \hline
LCM                      & 3.47             & \multicolumn{1}{c|}{0.45}                               & \multicolumn{1}{c|}{0.48}                               & 0.43                               \\ \hline
MoCaNet                  & 11.62            & \multicolumn{1}{c|}{0.84}                               & \multicolumn{1}{c|}{0.49}                               & 0.16                               \\ \hline
Ours                     & \textbf{17.68}   & \multicolumn{1}{c|}{\textbf{0.15}}                      & \multicolumn{1}{c|}{\textbf{0.25}}                      & \textbf{0.06}                      \\ \hline
Raw                      & 40.09            & \multicolumn{3}{c}{-}                                                                                                                                  \\ \hline
\end{tabular}%
}
\end{table}

\subsubsection{Evaluation of motion transfer quality}

Evaluating motion transfer quality remains challenging due to the absence of ground-truth motion clips in most datasets. In this paper, we utilize our proposed \textit{ID-Score} metric to quantitatively evaluate the extent of $IM$ information preservation of synthetic motion clips, thereby reflecting the performance of motion transfer. Therefore, we employ three gait recognition models as identity information extractors to calculate the \textit{ID-Score} on three datasets, as presented in Table~\ref{tab:idscore-tab}. As we can see, our method significantly outperforms the previous three methods with three gait recognition models across all datasets. This improvement is due to the fact that previous methods only consider $HS$ information, but \textit{IDPres} takes into account $IM$ information in addition to $HS$ information.

\begin{figure}[t]
    \centering
        \includegraphics[width=\linewidth]{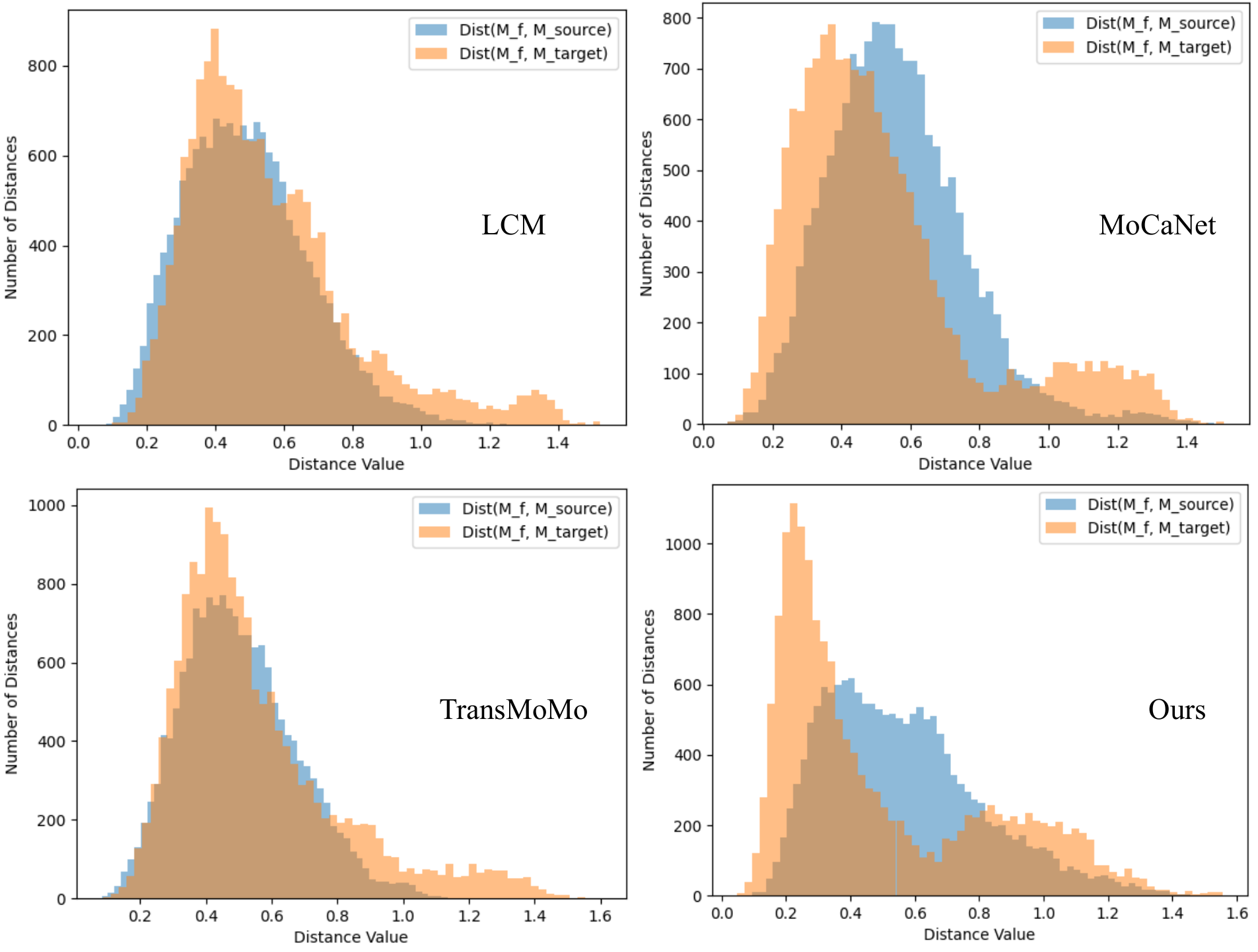}
    \caption{The identity histogram of different methods. $Dist(M_f, M_{source})$ (orange) and $Dist(M_f, M_{target})$ (blue) are calculated using the Euclidean metric within the identity space, representing how closely the generated motion $M_{f}$ distance with the source $M_{source}$ the target $M_{target}$ respectively. The x-axis means the distance values, and the y-axis means the number of distance values.}
    \label{fig:his_map}
\end{figure}

\begin{figure*}[t]
	\centering
	\includegraphics[width=0.9\textwidth]{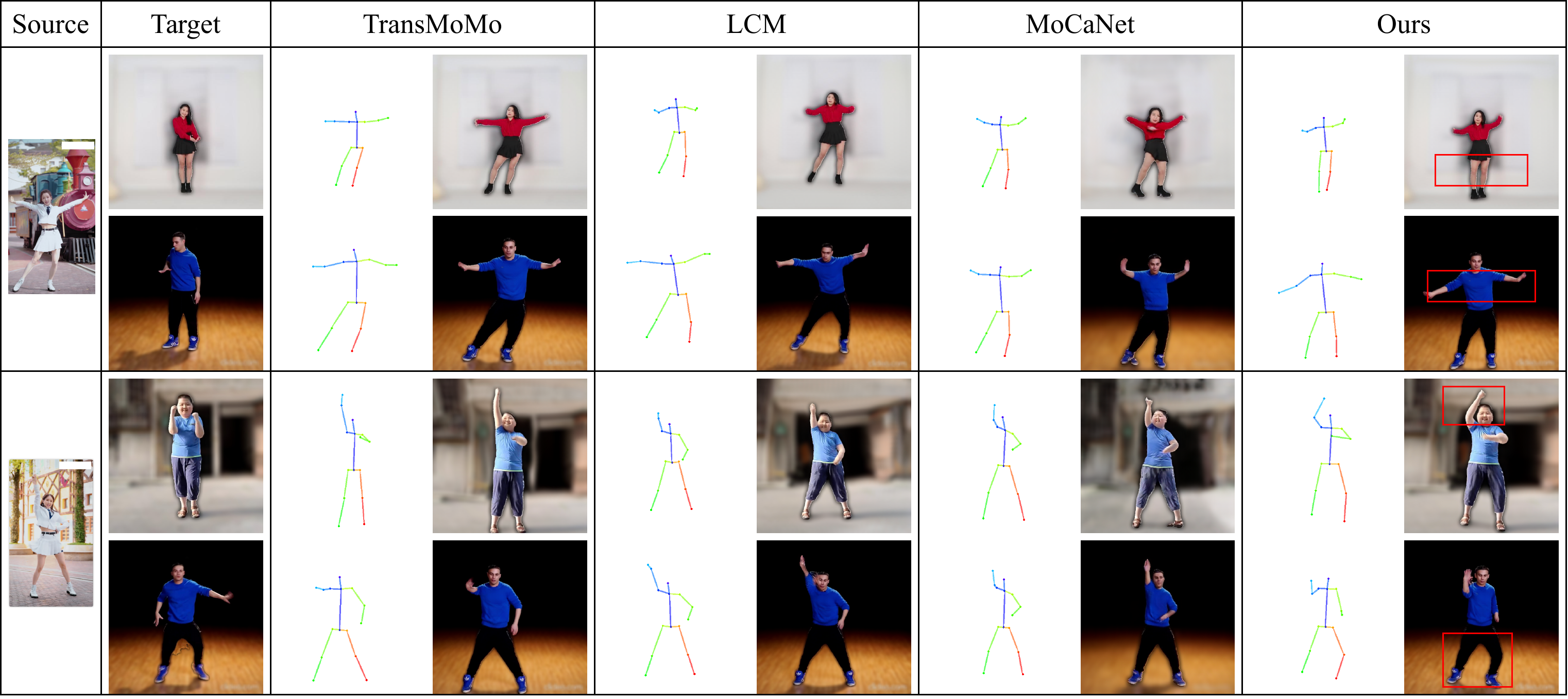}
	\caption{The rendered results of quantitative comparisons. Red boxes are the main differences between our method and other methods. Our proposed \textit{IDPres} can make the target person follow the source person to dance, but with his/her own style.}
	\label{fig:render_vis_com}
\end{figure*}

\subsubsection{Evaluation of the ability of cross-domain}
To quantitatively evaluate the cross-domain capabilities of our method and competitive methods, we adopt $Dancer99$ as the evaluation dataset. This dataset, distinct in choreography and subjects, differs from $Dancer101$, ensuring a robust evaluation of our method's adaptability to different domains. Specifically, we train all methods on the $Dancer101$ dataset and conduct testing on $Dancer99$. The metrics include $Rank_1^{rec}$, MSE, MAE, and FMD, and the results as shown in Table~\ref{tab:in-the-wild}. It is evident that \textit{IDPres} surpasses all competitive methods on four evaluation metrics. Notably, our method outperforms MoCaNet by 6.06\% in $Rank_1^{rec}$, keeping consistent with results in Table~\ref{tab:idscore-tab}. These results further demonstrate the superiority of our method in both $IM$ information preservation and motion reconstruction from a cross-domain perspective.

\subsubsection{Visualization of identity histogram}

In order to visualize the identity preservation performance of our method, we use Euclidean distances to measure the similarity between the generated motion clips and the source motion clips, as well as the generated motion clips and the target motion clips, as shown in Fig.~\ref{fig:his_map}. Regarding Euclidean distance histograms of different methods in Fig.~\ref{fig:his_map}, we first randomly sample 128 motion clips from the testing set of $Dancer101$ ($M_{source}$) and a corresponding in-the-wild dataset ($M_{target}$), respectively. Subsequently, using the $cross$ pipeline depicted in Fig.~\ref{fig:id_score} to obtain generated motion clip $M_{f}$. The Euclidean distances are then computed between $M_f$ and $M_{source}$/$M_{target}$ to indicate the extent of $IM$ information preservation.
The Euclidean distance distribution peaks, situated closer to the y-axis (more left axis), indicate that the synthetic motion clips have a high similarity to the motion clips they’re compared with.
As we can see, the distribution peak of our proposed method is around 0.25, outperforming those from LCM, MoCaNet, and TransMoMo, which stand at 0.4, 0.35, and 0.4, respectively.
Hence, the results of Euclidean distance distribution show that the synthetic motion clips of our proposed method preserve more information about the target’s identity information compared to previous methods.
It proves our method takes full use of $IM$ information when conducting the HMT task.

\subsubsection{Results of rendered videos}
In Fig.~\ref{fig:render_vis_com}, we further present the rendered video results of different methods in the wild videos. The first two columns indicate the source and target video clips and the first column of each method means the synthetic motion clips. Our method achieves better $IM$ performance than competitive methods. These results mean that the target subject can follow the source subject's dance content, but strongly with her/his own style. 
These findings demonstrate our proposed method can effectively capture the $IM$ information from the dance movements of target subjects, addressing the unnatural issue mentioned in Fig.~\ref{fig:challenges}.


\begin{figure*}[t]
	\centering
	\includegraphics[width=0.9\linewidth]{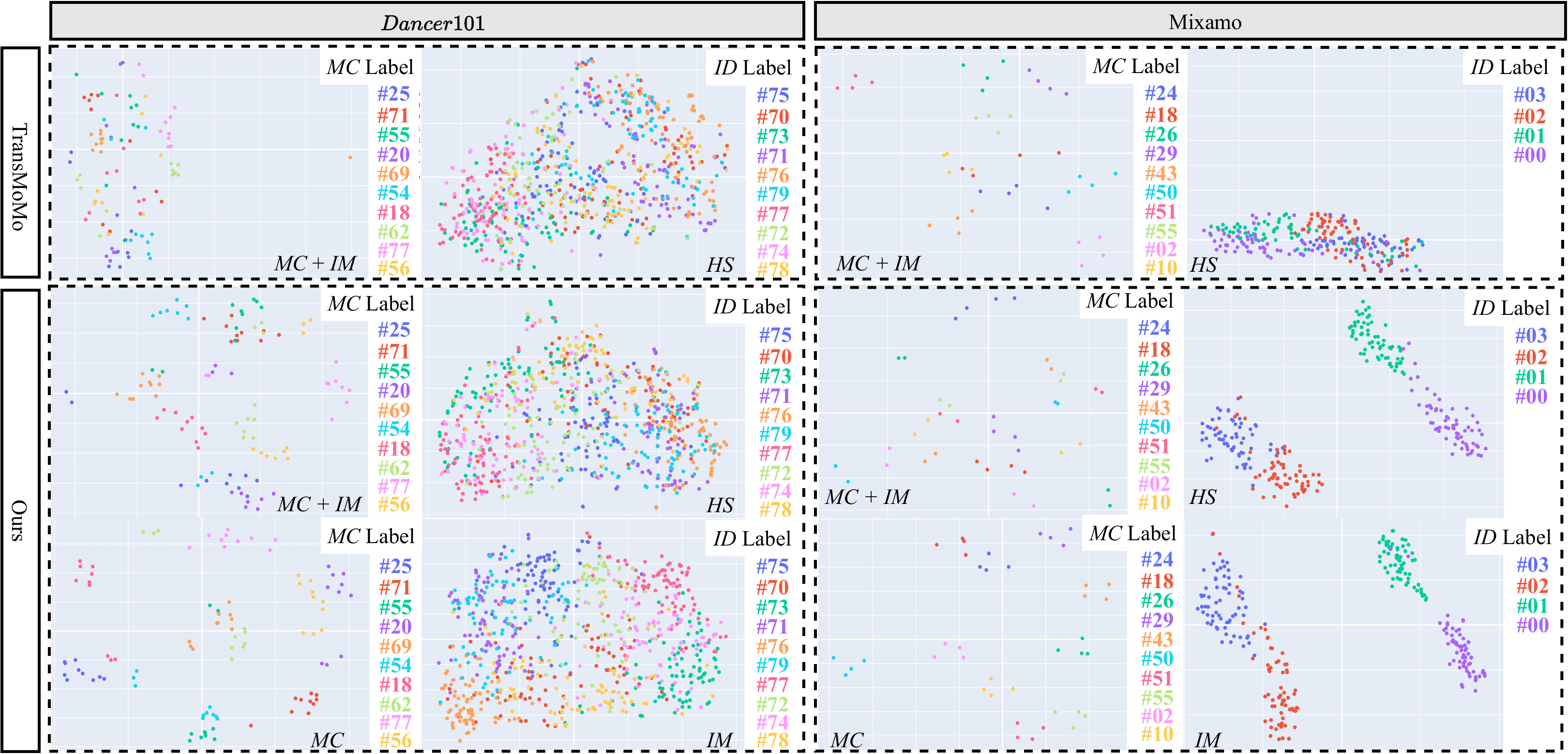}
	\caption{The results of representations in latent space using t-SNE for test datasets of $Dancer101$ and Mixamo. $MC$ $+$ $IM$ means concatenate $MC$ and $IM$ in channel axis before t-SNE. In addition, all results are shown for ten labels only.}
	\label{fig:representations}
\end{figure*}

\subsection{Performance comparison of different disentanglement combinations}
In this experiment, we generate videos by the permutation of $IM$, $MC$, and $HS$ representations.
Our primary aim is to evaluate how these different attributes contribute to a subject's identity information. 
We employ recognition accuracy as our evaluation metric, with the results for different attribute combinations on the $Dancer101$ dataset detailed in Table~\ref{tab:diff_input_frames}. 
GaitGraph~\cite{teepe2021gaitgraph} is utilized as the $\mathbf{E}_{gait}$ to extract identity information for all motion clips.
The results reveal that the transfer of different attributes impacts recognition accuracy in varying ways. 
These findings prove that our method can effectively disentangle and reassemble the three attributes. 
Notably, accuracy substantially decreases when preserving the target subject's $IM$ or $HS$, whereas preserving the target subject's $MC$ only slightly reduces accuracy, from 46.53\% to 35.31\%. Interestingly, we observed that preserving the target subject's $IM$ yields better recognition performance than preserving $HS$, suggesting that the human skeleton is more influential than individualized motion in identity preservation, a conclusion also echoed in other gait recognition studies~\cite{Lin2021ICCV, fan2023opengait}. This observation is further substantiated by the visualizations in Fig.~\ref{fig:diff_stage} $Exp_3$.

\begin{table}[t]
\caption{The rank 1 accuracies of gait recognition on $Dancer101$. Here 'A' means the source, and 'B' means the target. \ding{51} means the current attribute is from the corresponding subject (A or B). $M_f^{ABA}$ means  $MC$ and $HS$ are from the source (A), and $IM$ is from the target (B).}
\label{tab:diff_input_frames}
\centering
\resizebox{0.8 \linewidth}{!}{%
\begin{tabular}{c|c|c|c|c|c}
\hline
Aliases                      & \begin{tabular}[c]{@{}c@{}}Input \\ Motions\end{tabular} & $MC$         & $IM$ & $HS$       & Rank\_1 (\%)           \\ \hline\hline
\multirow{2}{*}{$M_f^{rec}$} ($\uparrow$) & A                                                        & \ding{51} & \ding{51}  & \ding{51} & \multirow{2}{*}{46.53} \\ \cline{2-5}
                             & B                                                        &            &             &            &                        \\ \hline
\multirow{2}{*}{$M_f^{ABA}$} ($\downarrow$) & A                                                        & \ding{51} &             & \ding{51} & \multirow{2}{*}{25.71} \\ \cline{2-5}
                             & B                                                        &            & \ding{51}  &            &                        \\ \hline
\multirow{2}{*}{$M_f^{ABB}$} ($\downarrow$) & A                                                        & \ding{51} &             &            & \multirow{2}{*}{10.61} \\ \cline{2-5}
                             & B                                                        &            & \ding{51}  & \ding{51} &                        \\ \hline
\multirow{2}{*}{$M_f^{AAB}$} ($\downarrow$) & A                                                        & \ding{51} & \ding{51}  &            & \multirow{2}{*}{14.29} \\ \cline{2-5}
                             & B                                                        &            &             & \ding{51} &                        \\ \hline
\multirow{2}{*}{$M_f^{BAA}$} ($\uparrow$) & A                                                        &            & \ding{51}  & \ding{51} & \multirow{2}{*}{35.31} \\ \cline{2-5}
                             & B                                                        & \ding{51} &             &            &                        \\ \hline
\end{tabular}%
}
\end{table}

\subsection{Disentanglement visualization}

Feature visualization can give an intuitive and straightforward impression of the performance of representation disentanglement. We try to compare the proposed \textit{IDPres} and TransMoMo method~\cite{yang2020transmomo} by visualization. 
The $MC$, $IM$, and $HS$ features are extracted by the disentanglement block $\mathbf{E}$.
Then, t-SNE~\cite{van2008visualizing} is used to project the disentangled features into a 2D space. The results are from different subjects (in different colours) by TransMoMo and our proposed \textit{IDPres}) are shown in Fig.~\ref{fig:representations}. 
The TransMoMo method, can not disentangle $IM$ representation from human movements. 
So, we use $MC$ $+$ $IM$ to represent the human movements. 
In order to provide fair comparisons, we provided 4 kinds of disentangled representation by \textit{IDPres}. They are $MC$ $+$ $IM$, $MC$, $IM$ and $HS$.

According to Fig.~\ref{fig:representations}, we can find the $MC$ $+$ $IM$ and $HS$ results tend to gather as a cluster if they are from the same category (the same colour in the figure) on both datasets. The results from the two methods all follow this rule. We can also find the $IM$ results by \textit{IDPres} gathered into different clusters on both datasets. This phenomenon demonstrates that the disentanglement block of \textit{IDPres} can effectively decompose the $IM$ representation. Moreover, the clusters on the Mixamo dataset are tighter than those on the $Dancer101$. The reason should be that data from Mixamo is virtual 3D animations, and not many variations are in the data. The data from $Dancer101$ was collected by cameras in real scenarios. 


\subsection{Ablation study}

We train different models to study the impact of different modules, such as triplet loss in disentanglement block, TPP layer in $E_{IM}$, learn-able identity weight in $E_{IM}$, the methods of different fusion, and different training schemes. 

\subsubsection{Effectiveness of triplet loss} 
\begin{figure}[t]
	\centering
	    \includegraphics[width=0.9\linewidth]{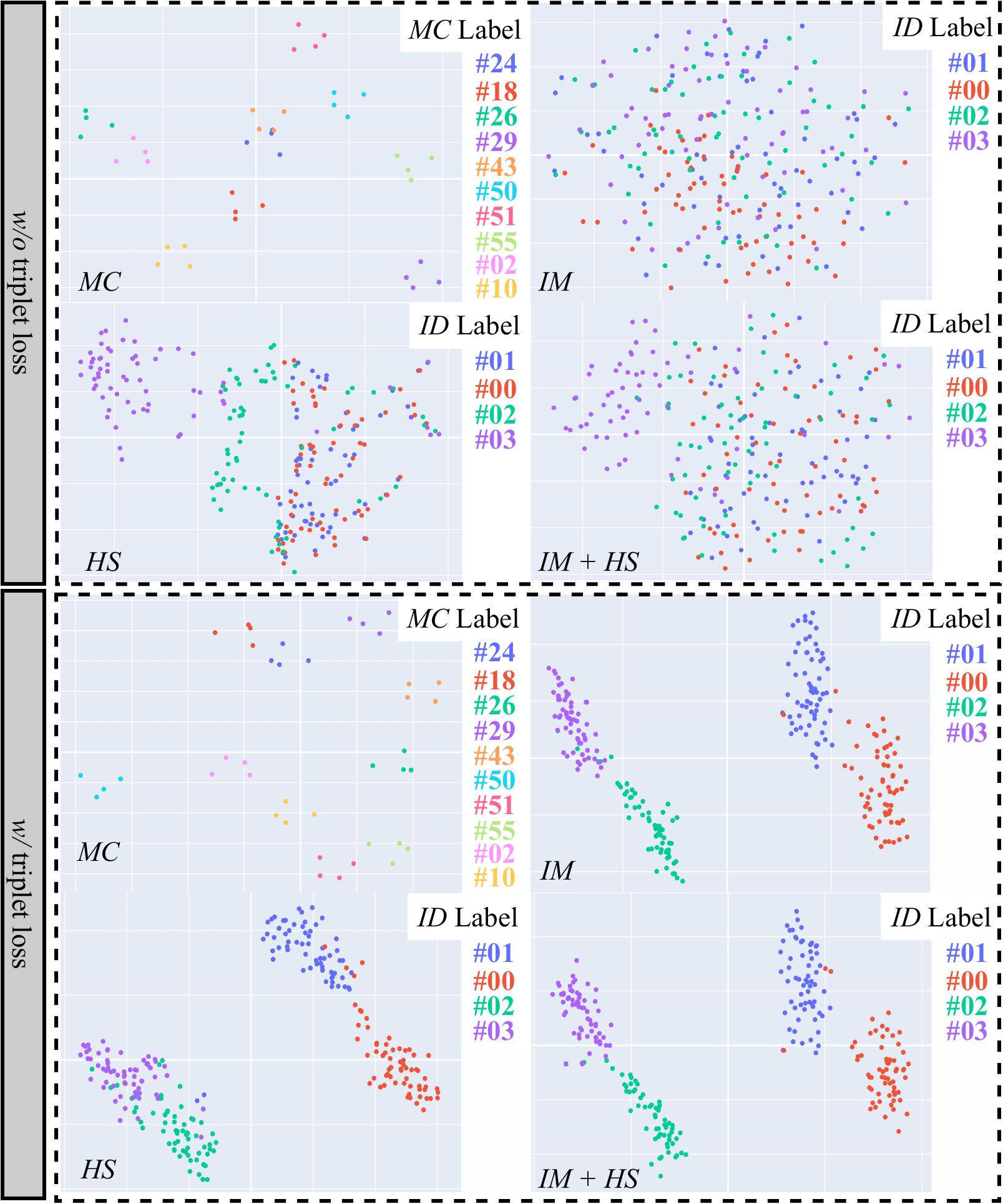}
	\caption{The ablation study of triplet loss on Mixamo dataset. Both $IM$ and $HS$ information ($IM$ + $HS$) typically represent the identity information of a subject.}
	\label{fig:w/o trip}
\end{figure}

It is known that triplet loss aims to pull intra-class samples close to each other while pushing inter-class samples away from each other~\cite{hermans2017defense}. To verify the effectiveness of the employed triplet loss, we train two model variants on the Mixamo dataset: \textit{w} triplet loss and \textit{w/o} triplet loss. Fig.~\ref{fig:w/o trip} shows the distribution in latent space as mentioned in three attributes. The result shows that triplet loss plays a vital role in decomposing different attributes.

\subsubsection{Effectiveness of TPP layer and learnable identity weight ($W_{id}$)}
Our TPP layer and the learnable identity weight are responsible for extracting the local details and picking individualized-related information from the many local textures, respectively. Using the same settings as the experiments in Table~\ref{tab:diff_input_frames}, Table~\ref{tab:ab_cops} shows the recognition accuracies of different synthetic motions by composing at different feedback levels in $E_{IM}$. The recognition accuracy will increase as the TPP layer and the learnable weight are adopted. This proves that the TPP layer and the learnable identity weight have positive effects on extracting the $IM$ representation in motion.

\begin{table*}[t]
\centering
\caption{Ablation experiments for our \textit{IDPres} model on $Dancer101$ dataset.}
\label{tab:ab_cops}
\resizebox{0.65\linewidth}{!}{%
\begin{tabular}{c|c|c|c|c|c} 
\hline
\multirow{3}{*}{Setting}                   & \multicolumn{5}{c}{Performances}                                                   \\ 
\cline{2-6}
                                           & $M_f^{rec}$ ($\uparrow$)      & $M_f^{ABA}$ ($\downarrow$)      & $M_f^{ABB}$ ($\downarrow$)     & $M_f^{AAB}$ ($\downarrow$)      & $M_f^{BAA}$ ($\uparrow$)      \\ 
\cline{2-6}
                                           & Rank1(\%)      & Rank1(\%)      & Rank1(\%)     & Rank1(\%)      & Rank1(\%)       \\ 
\hline
\textit{w/o} TPP and \textit{w/o} $W_{id}$ & 24.08          & 20.0           & \textbf{9.18} & \textbf{10.61} & 23.47           \\
\textit{w/o} $W_{id}$                      & 30.41          & \textbf{16.33} & 10.00         & 12.24          & 28.16           \\
\textit{w/o} TPP                           & 42.86          & 30.20          & 10.20         & 15.10          & 33.88           \\
Our full model                             & \textbf{46.53} & 25.71          & 10.61         & 14.29          & \textbf{35.31}  \\ 
\hline
\textit{w/o} AdaIN                         & 27.96          & 13.47          & 10.00         & 11.02          & 25.31           \\
\hline
\end{tabular}
}
\end{table*}

\subsubsection{Effectiveness of different fusion methods} 
In the proposed method, we use AdaIN in \textit{IBup} block to fuse $IM$ and $MC$ representations. The \textit{w/o} AdaIN means not using AdaIN and using a concatenate operator on the channel axis for $IM$ and $MC$. Quantitative results of the different metrics are reported in Table~\ref{tab:ab_cops}. The results show that using AdaIN outperforms concatenate fusion settings. Since AdaIN operations after each convolution layer of the \textit{IBup} block can modify the second-order statistics of $IM$ representations and act on different motion scales. 

\begin{figure}[t]
    \centering
        \includegraphics[width=0.9\linewidth,height=6.5cm]{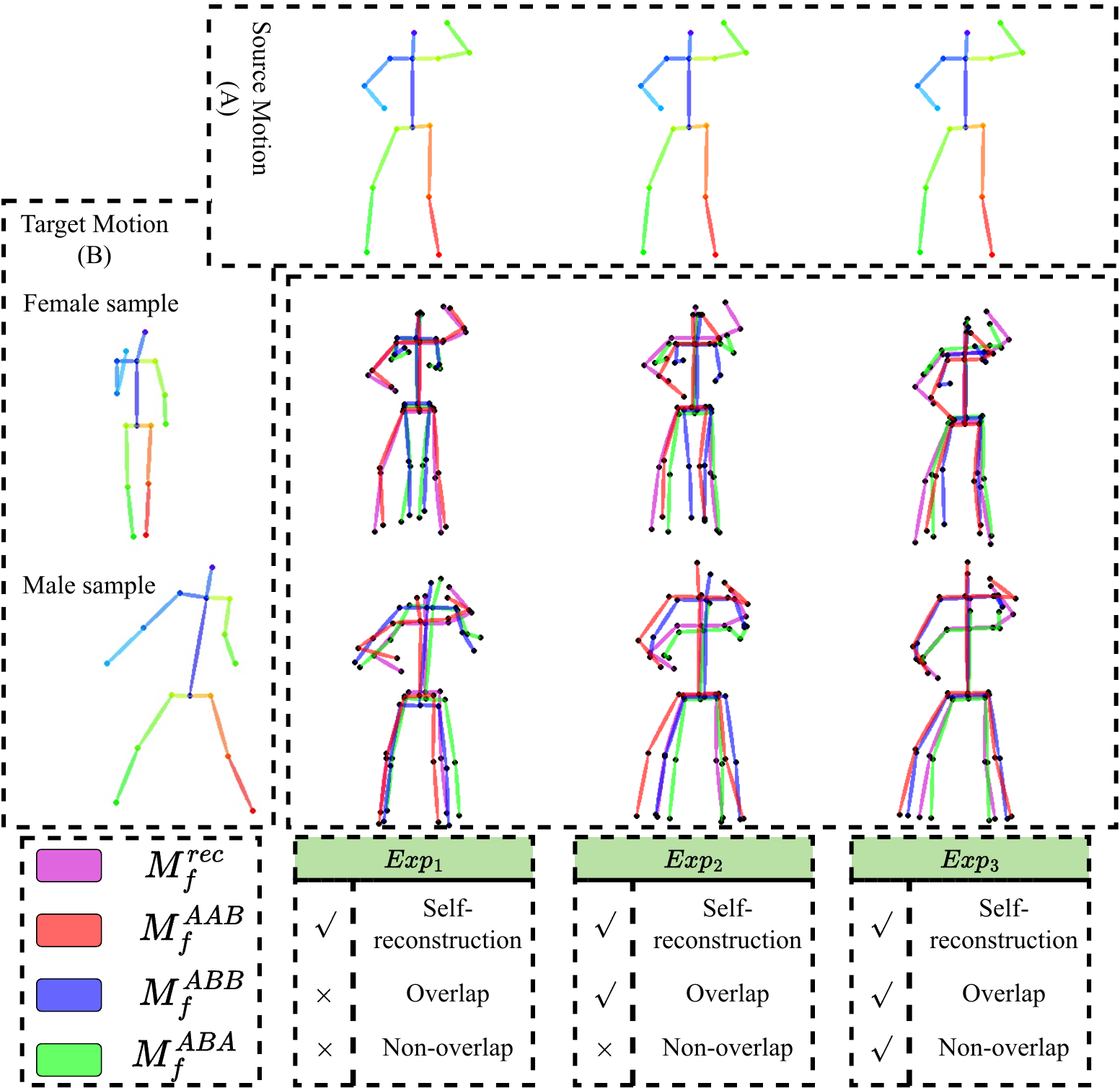}
    \caption{Ablation experiments for different training schemes.}
    \label{fig:diff_stage}
\end{figure}

\subsubsection{Effectiveness of different training schemes} 
Some examples of different training schemes are shown in Fig.~\ref{fig:diff_stage}. In Experiment $Exp_1$,  the generated motion $M^{ABB}_{f}$ (blue motion) still preserves the $IM$ of the target subject. It can be seen that as training steps increase, the quality of the generated motion $M^{ABB}_{f}$ has positive effects. The results show the benefits of adopting the transferring motion training schemes (overlap and non-overlap) over only the self-reconstruction. Our designed training schemes provide powerful information swapping for the disentanglement block and the generator, which allows us to effectively disentangle different attributes and then map them correctly to the corresponding input motion. 




\section{Conclusion and Future Work}

In this paper, we introduce an innovative identity-preserved human motion transfer method, \textit{IDPres}, capable of generating realistic human motion videos while preserving the target subject's identity information.
The key to achieving realistic motion transfer is the consideration of $IM$ information. 
To effectively disentangle $IM$ from human movements,  \textit{IDPres} contains three efforts: the TPP and learnable identity weight, the implementation of a triplet loss function, and the application of overlap and non-overlap training schemes. 
Additionally, thanks to the success of gait recognition methods in capturing identity information, we leverage gait models as identity information extractors and design a novel metric, \textit{ID-Score}, to evaluate the effectiveness of both $IM$ information disentanglement.
Considering the absence of public datasets with paired identity-movement data and complex motion, we collect a dataset, $Dancer101$, where different subjects perform identical dance motions, facilitating the study of identity-preserved HMT. 
Our experimental findings demonstrate that our method can generate natural human motion videos, enabling the target subject to imitate the source subject's motion while maintaining more identity information.

HMT is a very challenging research topic because the data is in the spatio-temporal domain. There are still no effective HMT methods to generate a high-resolution and realistic long video. The proposed method involves identity preservation to improve the movement's quality. To continuously improve HMT, training on a large dataset is needed. The data collection and labelling are difficult and time-consuming. A possible solution is contrastive learning with unlabeled data. To improve the quality, another direction is to disentangle more attributes, including gender, age, clothing, etc. A fine-grained description of human motion can be obtained by a lot of attributes, and the quality of HMT will also be improved.

\noindent \textbf{Privacy concerns.} The datasets $Dancer101$ and $Dancer99$ were collected in accordance with Standard both YouTube and Bilibili License.
The release of these datasets is executed through the provision of links to videos on YouTube and Bilibili, and pre-processed skeleton points data, fully following good practice of similar data collection~\cite{fan2023exploring, sevastopolskiy2023boost, yang2020transmomo}.

\section*{Acknowledgments}
We greatly thank Mr. Chao Fan for his help and valuable suggestions, significantly improving the quality of our paper.
This work was supported 
in part by the National Natural Science Foundation of China under Grant 61976144, 
in part by the National Key Research and Development Program of China under Grant 2020AAA0140002, 
and in part by the Shenzhen Technology Plan Program (Grant No. KQTD20170331093217368).


\bibliographystyle{IEEEtran}
\bibliography{tifsbib}


\begin{IEEEbiography}[{\includegraphics[width=1in,height=1.25in,clip,keepaspectratio]{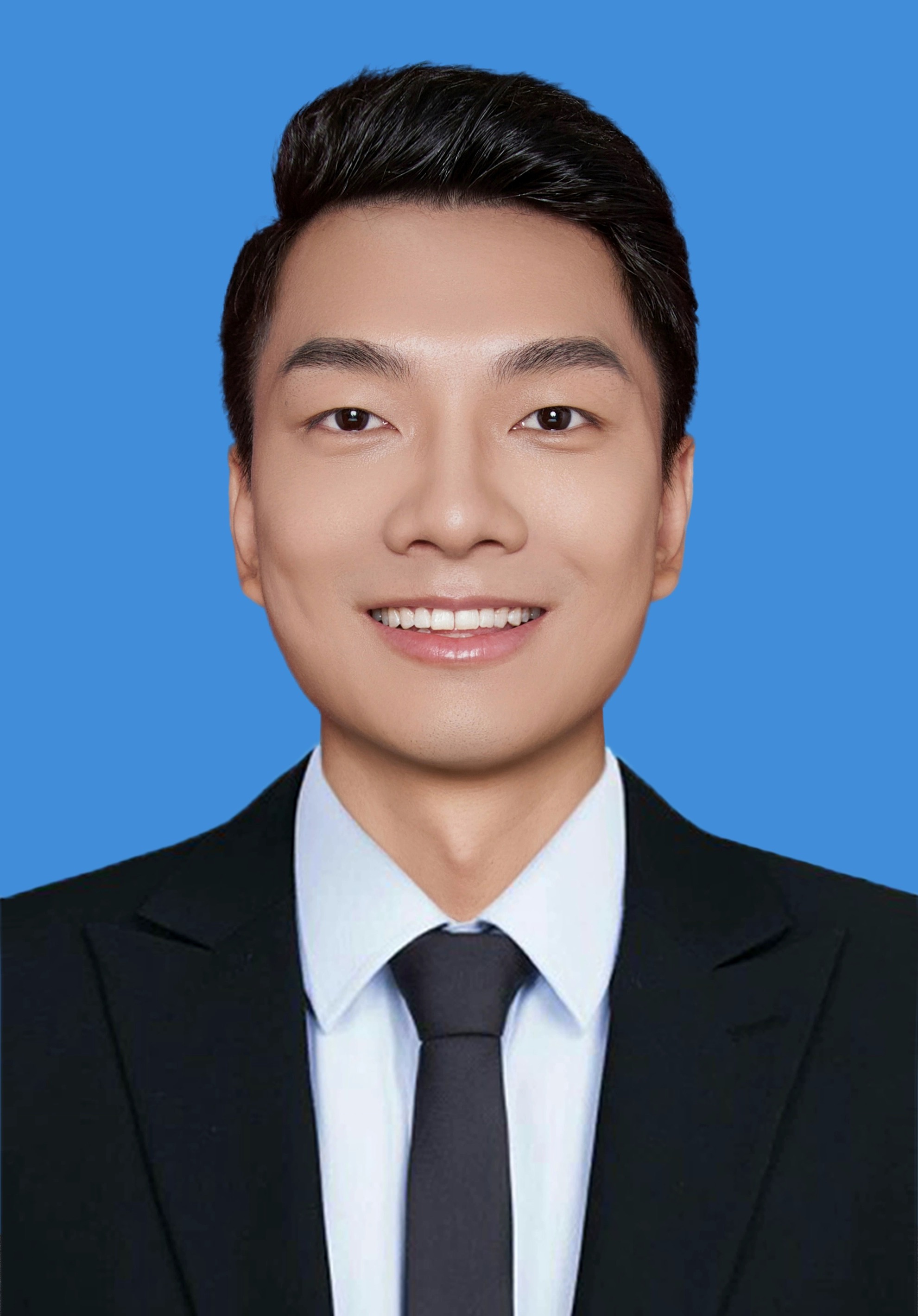}}]{Jingzhe Ma}
received the B.E. and M.S. degrees in computer science and technology from Zhengzhou University in 2017 and 2020, respectively. He is currently a Ph.D. candidate with Department of Computer Science and Engineering, Southern University of Science and Technology. His research interests include human video synthesis and gait recognition.
\end{IEEEbiography}
\begin{IEEEbiography}[{\includegraphics[width=1in,height=1.25in,clip,keepaspectratio]{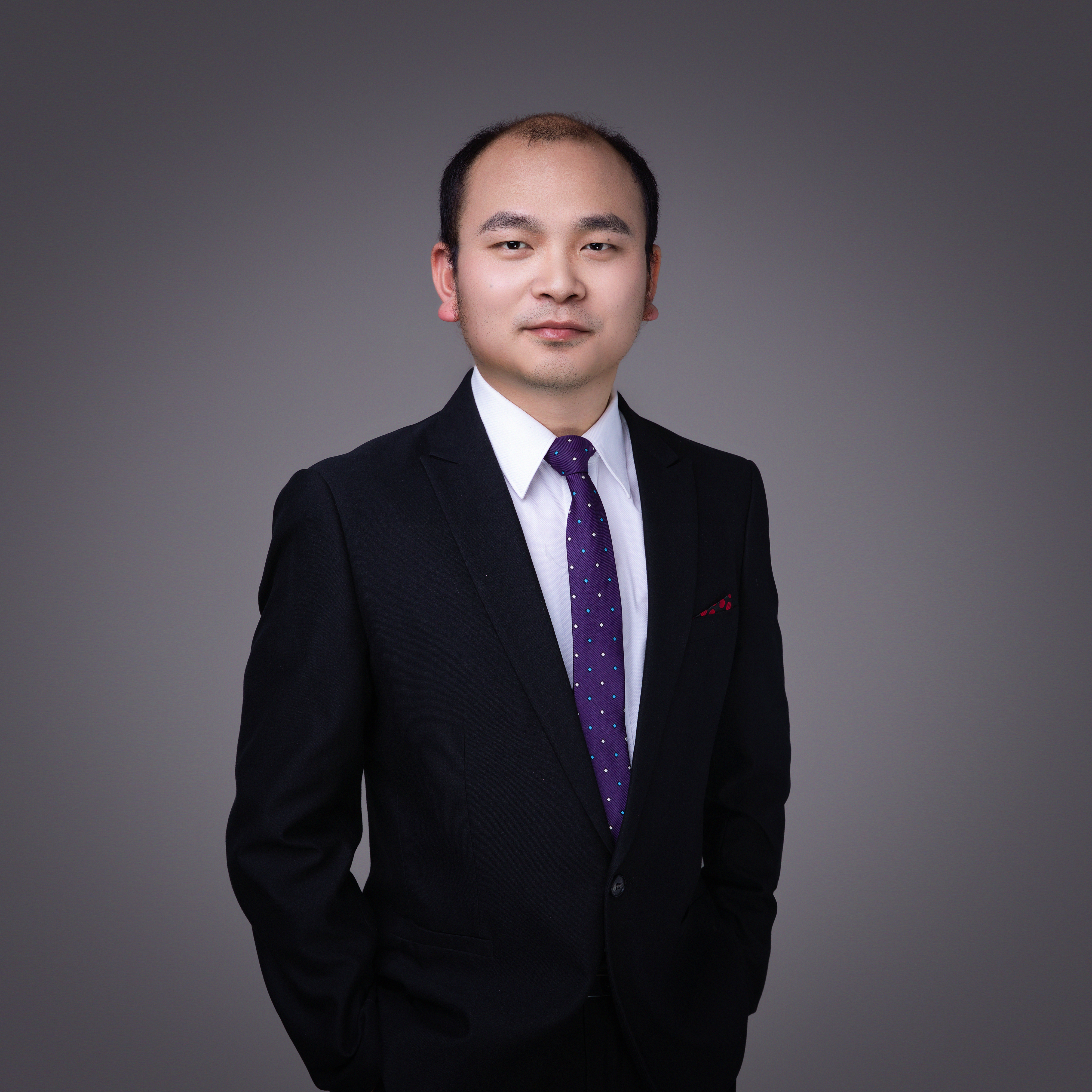}}]{Xiaoqing Zhang}
received his B.S. degree in water conservancy and hydropower engineering from South China Agricultural University in 2016, M.S. degree in computer technology from Zhengzhou University in 2019, and Ph.D. degree in mechanics from Southern University of Science and Technology (SUSTech) in 2023. He is currently a visiting scholar in Department of Computer Science and Engineering, SUSTech. His research interests include feature disentanglement, interpretability, and medical image processing.
\end{IEEEbiography}
\begin{IEEEbiography}[{\includegraphics[width=1in,height=1.25in,clip,keepaspectratio]{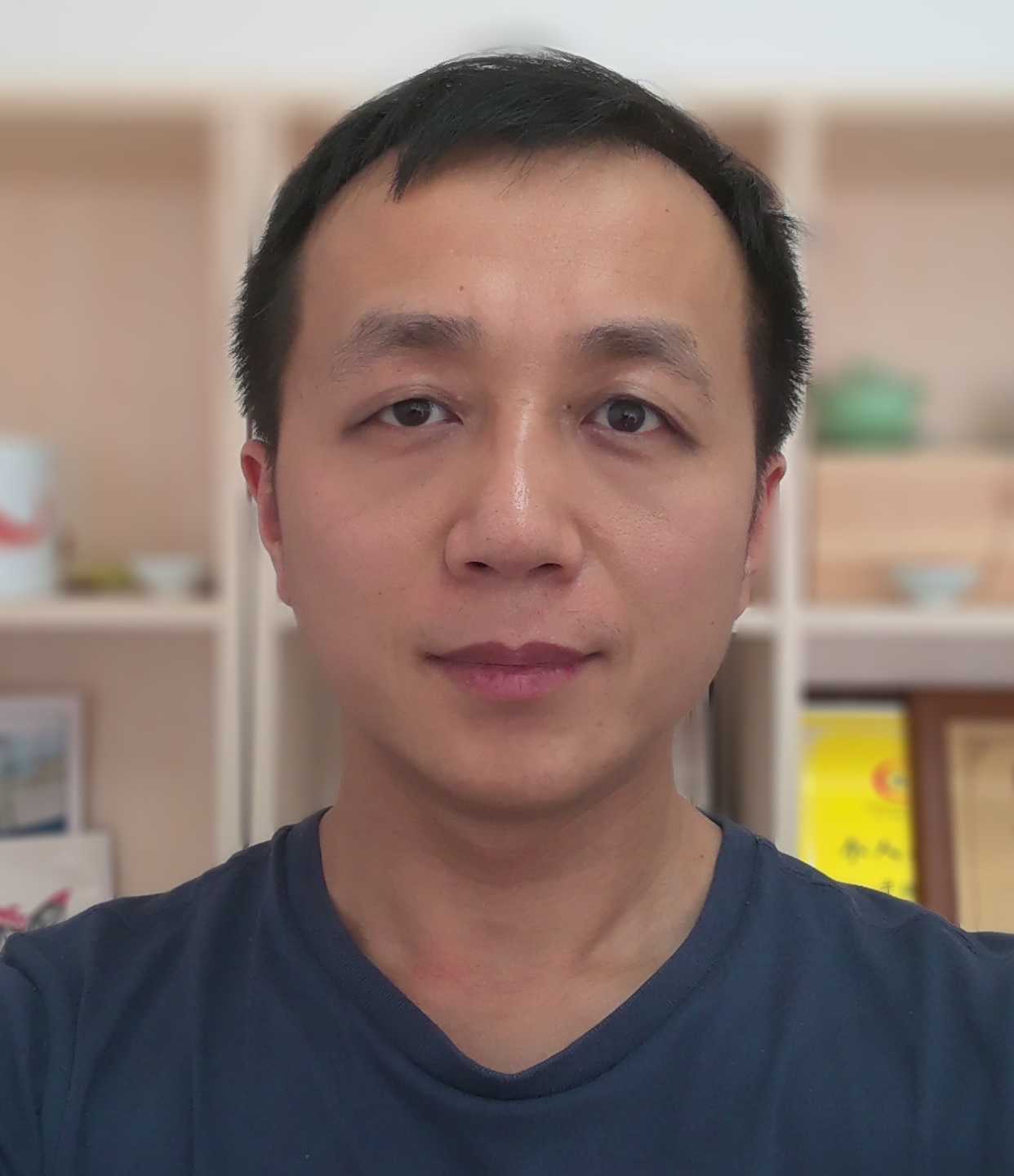}}]{Shiqi Yu} is currently an Associate Professor in the Department of Computer Science and Engineering, Southern University of Science and Technology, China. He received his B.E. degree in computer science and engineering from the Chu Kochen Honors College, Zhejiang University in 2002, and Ph.D. degree in pattern recognition and intelligent systems from the Institute of Automation, Chinese Academy of Sciences in 2007. He worked as an Assistant Professor and an Associate Professor at Shenzhen Institutes of Advanced Technology, Chinese Academy of Sciences from 2007 to 2010, and as an Associate Professor at Shenzhen University from 2010 to 2019. His research interests include gait recognition, face detection and computer vision.
\end{IEEEbiography}

\vfill\eject

\end{document}